\documentclass[twoside]{article}

\usepackage{amsmath}
\usepackage{amsfonts}
\usepackage{mathtools}
\usepackage{algpseudocode}
\usepackage{algorithm}
\usepackage{varwidth}
\usepackage{subcaption}

\usepackage{amsthm}
\usepackage{lmodern}

\usepackage{layouts}

\usepackage{tikz}
\usepackage{tikz-cd}
\usepackage{pgfplots}

\usepackage{xcolor}
\usepackage{soul}

\newtheorem{theorem}{Theorem}
\newtheorem{corollary}{Corollary}

\usepackage{hyperref}

\usepackage[accepted]{aistats2025}
\usepackage[round,authoryear]{natbib}

\usepackage{subfiles} 

\pgfplotsset{compat=1.18}

\begin{document}


%

%

\twocolumn[

\aistatstitle{Estimating the Spectral Moments of the Kernel Integral Operator from Finite Sample Matrices}

\aistatsauthor{ Chanwoo Chun\textsuperscript{1,4} \And SueYeon Chung\textsuperscript{2,4} \And Daniel D. Lee \textsuperscript{3,4}}

\aistatsaddress{ \textsuperscript{1} Weill Cornell \\Medical College \And  \textsuperscript{2} New York University \And  \textsuperscript{3} Cornell Tech \And \textsuperscript{4} Flatiron Institute} ]

\begin{abstract}
Analyzing the structure of sampled features from an input data distribution is challenging when constrained by limited measurements in both the number of inputs and features.
Traditional approaches often rely on the eigenvalue spectrum of the sample covariance matrix derived from finite measurement matrices; however, these spectra are sensitive to the size of the measurement matrix, leading to biased insights. In this paper, we introduce a novel algorithm that provides unbiased estimates of the spectral moments of the kernel integral operator in the limit of infinite inputs and features from finitely sampled measurement matrices. Our method, based on dynamic programming, is efficient and capable of estimating the moments of the operator spectrum. We demonstrate the accuracy of our estimator on radial basis function (RBF) kernels, highlighting its consistency with the theoretical spectra.
Furthermore, we showcase the practical utility and robustness of our method in understanding the geometry of learned representations in neural networks.
\end{abstract}

\section{INTRODUCTION}
A primary objective of statistical inference in machine learning is to accurately estimate the characteristics of a high-dimensional distribution based on finite samples. For example, consider a Gaussian process with an unknown covariance where only a limited set of sample functions is drawn from the process. In many cases, we cannot observe the functions themselves, but rather their noisy evaluations at sampled input points \citep{williams2006gaussian}. What can we infer about the underlying process from such a sampled set of functions and input points? Similarly, in large-scale neural networks, we aim to understand the characteristics of the neural feature representations as both the number of features and input points grow infinitely large \citep{cho2009kernel,mei2018mean,chung2018classification,cohen2020separability,canatar2021spectral,canatar2024spectral}.  The central question of this work is how to estimate the spectral properties of the underlying infinite process when both features and evaluation points are finitely sampled. 

A matrix can be constructed from sampled measurements of the process, where each row corresponds to an individual input sample and each column is a sampled feature. It is common practice to analyze the eigenvalue spectrum of the sample covariance matrix derived from this finite measurement matrix. However, this spectrum is biased, leading to inaccurate insights into the underlying structure. Therefore, prior work aims to correct this bias under the assumption that the rows are sampled \citep{kong2017}. However, in the setup where both rows and columns are sampled, this method produces biased estimates.

In our model, we consider the measurement matrix as a sampled submatrix of a larger underlying matrix where both the number of rows and columns approach infinity. A kernel integral operator can be defined as the expected covariance of this larger matrix and we study how to accurately infer the spectral properties of this operator, in particular its spectral moments.
We propose a novel, computationally efficient algorithm based upon dynamic programming, to estimate the spectral moments of the underlying kernel operator from a finite measurement matrix.

In the following, we first describe a mathematical framework relating a finite measurement matrix to the spectrum of a kernel integral operator \citep{bach2017equivalence}.  We show how the naive estimators of the spectral moments based upon the finite covariance matrix are biased.  Then we introduce our method for estimating the spectral moments by averaging appropriate products of non-repeating cycles in the measurement matrix. Our method employs a recursive procedure that is computationally efficient, polynomial in the size of the matrix and the order of the moments.  We demonstrate the accuracy of our method with the radial basis function (RBF) kernel operator, where a direct comparison to the theoretical spectrum and to other estimation methods is possible. We also demonstrate inferring the eigenvalues of kernel integral operators from our moment estimates using an existing algorithm. Finally, we show how our estimates can be used to analyze the learning dynamics of a rectified linear unit (ReLU) neural network during feature learning, showing how networks of different widths can be related by their kernel operators.

\section{KERNEL OPERATOR}

\subsection{Kernel as expectation}

We model the entries of a $P\times Q$ measurement matrix $\left[\Phi_{i\alpha}\right]$ as arising from the following stochastic process. Each row $i\in\{1,...,P\}$ is characterized by a latent input variable $x_{i}$ drawn independently from a probability measure $\rho_{{\mathcal X}}(x)$ over a latent space $\mathcal{X}$, and each column $\alpha\in\{1,...,Q\}$ is characterized by a latent variable $w_{\alpha}$ drawn independently from a probability measure $\rho_{{\mathcal W}}(w)$ over a latent space $\mathcal{W}$. In random feature networks, for instance, $x_i$ and $w_\alpha$ can be seen as an input pattern and neural weights respectively.  The $(i\alpha)$-th entry of the matrix $\left[\Phi_{i\alpha}\right]$ is produced by a function $\phi$
that maps the pair $(x_{i},w_{\alpha})$ to a real number:
\begin{equation}
\Phi_{i\alpha}=\phi(x_{i},w_{\alpha}).
\end{equation}
A kernel function $k:{\mathcal X}\times{\mathcal X}\rightarrow\mathbb{R}$ over input pairs
can then be defined as the expected value of the product of $\phi$ over the features \citep{bach2017equivalence}:
\begin{equation}
k(x,x^{\prime})=\int d\rho_{\mathcal{W}}(w)\:\phi(x,w)\phi(x^{\prime},w).
\end{equation}
We assume the function $\phi$ is square-integrable with respect to
both $\rho_{{\mathcal X}}$ and $\rho_{{\mathcal W}}$, so the kernel function
is positive-definite and bounded. The tuple $(\phi,\rho_\mathcal{X},\rho_\mathcal{W})$ uniquely defines a generative process for the measurement matrices and corresponding kernel.

\subsection{Kernel integral operator}

Now consider the integral operator
$T_{k}:\mathcal{L}^{2}(\mathcal{X},\rho_{\mathcal{X}})\to \mathcal{L}^{2}(\mathcal{X},\rho_{\mathcal{X}})$ \citep{cucker2002mathematical}:
\begin{equation}
T_{k}f\coloneqq\int d\rho_{{\mathcal X}}(x)\:k(\cdot,x)f(x).
\end{equation}
Since $\phi$ is square-integrable, $T_{k}$ is a trace class operator.
Therefore $T_{k}$ is a compact, bounded, and self-adjoint linear
operator, whose spectrum consists of a countable, non-increasing sequence of eigenvalues $\left\{ \lambda_{l} \ge 0 \right\} _{l=1}^{\infty}$.
The corresponding eigenfunctions are defined implicitly as $\lambda_l e_l= T_k e_l$ where $e_l\in\mathcal{L}^2\left(\mathcal{X},\rho_\mathcal{X}\right)$ and $\left\{ e_{l}\right\} _{l=1}^{\infty}$ forms an orthonormal set. Since the product of a bounded operator and trace class
operator is also trace class, $\text{tr}T_{k}^{n}$ is well-defined
for any positive integer $n$. We define the spectral $n$-th moment $m(n)$ as the sum over the $n$-th powers of the eigenvalues:
\begin{equation}
m(n)\coloneqq\sum_{l=1}^{\infty}\lambda_{l}^{n}\equiv\text{tr}T_{k}^{n}.
\end{equation}
The moments can also be written as the expectation over a product of kernel functions,
\begin{equation}
m(n)=\int\prod_{j=1}^{n}d\rho_{{\mathcal X}}(x_{j})\:\prod_{j=1}^{n}k(x_{j},x_{j+1})
\end{equation}
with the constraint that $x_{n+1}=x_{1}$. The moments $\left\{m(n) \right\}_{n=1}^\infty$ uniquely determine the spectrum of $T_k$ via the Stieltjes transform (see Appendix). Various methods to estimate the spectral moments $m(n)$ of $T_k$ from a finite measurement matrix $\left[\Phi_{i\alpha}\right]$ are investigated in this work.

\subsection{Naive estimator}

Although the rows and columns of the measurement matrix $\Phi$ are sampled independently, the matrix coefficients will be correlated due to similarities in the sampled inputs and features. The conventional approach to analyze the spectral structure of the measurement matrix is to form the Gram matrix $K\in\mathbb{R}^{P\times P}$ that represents the similarity between input rows:
\begin{equation}
K_{ij}=\frac{1}{Q}\sum_{\alpha=1}^{Q}\phi(x_{i},w_{\alpha})\phi(x_{j},w_{\alpha})=\frac{1}{Q}\sum_{\alpha=1}^{Q}\Phi_{i\alpha}\Phi_{j\alpha}.
\end{equation}
The matrix $K$ is positive semi-definite and its moments are given by $\text{tr}K^{n}$.  If the similarity between different inputs $x_i$ and $x_j$ is ${\mathcal O}(1)$, then we expect the trace $\text{tr}K^{n}$ to scale as ${\mathcal O}(P^{n})$.  We normalize the traces by this scaling factor to give the naive estimator $\hat{m}_0(n)$:
\begin{equation}
\label{eq:naive}
\hat{m}_{0}(n)=\text{tr}\left[\left(\frac{K}{P}\right)^{n}\right].    
\end{equation}

In the limit of large $P$ and $Q$, the naive estimates 
 will converge to the moments of the kernel integral operator, $\hat{m}_{0}(n)\to m(n)$.
For $n=1$, $\hat{m}_{0}(1)$ is the sample variance of $\phi(x,w)$ and is an unbiased estimate of $m(1)$.
However, for $n>1$ and finite $P$ and $Q$, 
$\hat{m}_{0}(n)$ is a biased estimate
of $m(n)$.
To understand why $\hat{m}_{0}(n)$ is biased, consider the expected value of
the second moment estimate:
\begin{multline}\label{eq:naive2}
\left\langle \hat{m}_{0}(2)\right\rangle _\Phi = \frac{1}{P^{2}}\frac{1}{Q^{2}}\sum_{i,j}^{P}\sum_{\alpha,\beta}^{Q}\left\langle \Phi_{i\alpha}\Phi_{j\alpha}\Phi_{j\beta}\Phi_{i\beta}\right\rangle 
\\
=\frac{1}{P^{2}Q^{2}}\sum_{i\neq j}^{P}\sum_{\alpha\neq \beta}^{Q}\left\langle \Phi_{i\alpha}\Phi_{j\alpha}\Phi_{j\beta}\Phi_{i\beta}\right\rangle
\\
+\frac{1}{P^{2}Q^{2}}\sum_{i}^{P}\sum_{\alpha\neq \beta}^{Q}\left\langle \Phi_{i\alpha}^{2}\Phi_{i\beta}^{2}\right\rangle
+\frac{1}{P^{2}Q^{2}}\sum_{i\neq j}^{P}\sum_{\alpha}^{Q}\left\langle \Phi_{i\alpha}^{2}\Phi_{j\alpha}^{2}\right\rangle
\\
+\frac{1}{P^{2}Q^{2}}\sum_{i}^{P}\sum_{\alpha}^{Q}\left\langle \Phi_{i\alpha}^{4}\right\rangle .
\end{multline}
The second term in this expansion contains the connected product
$\left\langle \Phi_{i\alpha}^{2}\Phi_{i\beta}^{2}\right\rangle$ whose expected value is
$\int d\rho_{\mathcal{X}}(x)\:k(x,x)^2$, and differs from the second moment of the kernel integral operator.  This term is order ${\mathcal O}\left(\frac{1}{P}\right)$ and introduces a finite sampling bias into the estimate $\hat{m}_{0}(2)$.
Similarly, the third and fourth terms in the expansion will give rise to bias terms of order ${\mathcal O}\left(\frac{1}{Q}\right)$ and ${\mathcal O}\left(\frac{1}{PQ}\right)$ respectively.

This analysis generalizes to all higher moments $n>1$. The naive estimate $\hat{m}_{0}(n)$ derived from the sample Gram matrix contains biased terms of order 
${\mathcal O}\left(\frac{1}{P}+\frac{1}{Q}\right)$.

\section{RELATED WORK}

\subsection{Random matrix theory}

Random matrix theory analyzes the spectral characteristics of ensembles of Wishart matrices. The basic theory considers a Wishart matrix formed by taking the covariance of a large $P\times Q$ random matrix $\left[\Phi_{i\alpha}\right]$, whose entries are independently sampled from the standard normal distribution, i.e. $\Phi_{i\alpha}\sim{\mathcal N}(0,1)$. The spectrum of the Wishart matrix converges to a well-defined limit as $P,Q\rightarrow \infty$ when the ratio $\frac{P}{Q}$ is fixed.  This limiting spectral distribution differs from that of an identity covariance matrix and is known as the Marchenko-Pastur distribution \citep{marchenko1967distribution}.

Within our framework, a measurement matrix with independent and identically distributed (i.i.d.) normal entries is generated by taking $x,w\in\mathbb{R}^d$ with $\rho_{\mathcal{X}}=\mathcal{N}\left(0,I_{d\times d} \right)$, $\rho_{\mathcal{W}}=\mathcal{N}\left(0,\frac{1}{d}I_{d\times d} \right)$ and
bilinear map $\phi(x,w)=x^\top w$.  When $d$ approaches infinity and is much larger than $P$ and $Q$, each element $\Phi_{i\alpha}$ becomes an independent standard normal random variable.

In this case, the moments of the kernel integral operator are $m(n)=d^{-(n-1)}$. For large $d\gg P,Q$, the spectrum of the sample Gram matrix and the corresponding naive spectral moment estimates will be dominated by bias terms, and the leading order fully-connected bias terms are the same terms that give rise to the Marchenko-Pastur distribution. In the next section, we will see how to better estimate the spectral moments from measurement matrices by eliminating the bias terms.

\subsection{Estimator for fully observed features}

\cite{kong2017} consider the problem where the inputs are sampled from an underlying distribution but the features are fully observed with finite cardinality $d$, e.g. $w_\alpha \in \{w_1,w_2,...,w_d\}$. In their scenario, the measurements can be modeled as a $P\times d$ matrix $\bar{\Phi}\in\mathbb{R}^{P\times d}$; other work in the spectrum estimation literature also considers similar problem setup \citep{ledoit2004well,burda2004signal,el2008spectrum,khorunzhiy2008estimates,bhattacharjee2024sublinear}.
\cite{kong2017} models the observed measurement matrix as a matrix sketching process: $\bar{\Phi}_{i \alpha }=\sum_{k=1}^{d}x_{ik}F_{k\alpha}$ where the coefficients $x_{ik}$ are independently sampled from the standard normal distribution, and $F\in\mathbb{R}^{d\times d}$ is a deterministic matrix. Their method seeks to estimate the spectral moments of $S\coloneqq\frac{1}{d}FF^{\top}$, i.e. $m_\text{KV}(n)=\text{tr}\left(S^n\right)$ in order to obtain the spectrum of $S$.

Note that the naive estimator based upon the Gram matrix $\bar{K}\coloneqq\frac{1}{d}\bar{\Phi}\bar{\Phi}^{\top}$ with spectral moments
$\hat{m}_0(n) = \text{tr}(\hat{K}^n)$ is biased.  Instead, the following simple unbiased estimator is proposed:
\begin{equation}
\hat{m}'_{\text{KV}}(n)=\prod_{l=1}^{n}\bar{K}_{i_{l}i_{l+1}}
\end{equation}
where the product indices $i_{l}\in\left\{ 1,\ldots,P\right\} $
are disjoint, $i_{l}\neq i_{k}$ for all $l\neq k$, except for the trace constraint $i_{1}=i_{n+1}$. The proof that this simple estimator is unbiased relies on the assumption that $\bar{\Phi}_{i \alpha }$ is zero-mean, which is not a requirement for our model. With only a single realization of $\left\{ i_{l}\right\} _{l=1}^{n}$, $\hat{m}'_{\text{KV}}(n)$, the variance of the estimate is high. It would be optimal to average over all possible realizations of $\left\{ i_{l}\right\} _{l=1}^{n}$, but there is no known computationally efficient algorithm to perform the sum for large $n$. Thus, the authors propose averaging over sets of increasing indices, i.e. $i_{1}<i_{2}<\ldots<i_{n}$.  This leads to their estimator which considers the trace of the following matrix product:
\begin{equation}
    \hat{m}_\text{KV}(n)=\frac{\text{tr}\left( \bar{K}^{n-1}_\text{up} \bar{K} \right) }{\binom{P}{n}} .
\end{equation}
$\bar{K}_\text{up}$ is the upper triangular matrix formed from $\bar{K}$, e.g. the diagonal and lower triangular entries are set to zero.

The estimator $\hat{m}_\text{KV}(n)$ can be used with the measurement matrix $\Phi$ in our problem setup in two ways. 
One is setting $\bar{\Phi}\leftarrow \Phi$ which is equivalent to treating $\Phi$ as if all features are observed.  The other is to set $\bar{\Phi}\leftarrow \Phi^{\top}$, which is equivalent to the assumption that all the inputs are observed but the features are sampled.
We refer to the former estimator as $\hat{m}_\text{KV-row}(n)$ and the latter as $\hat{m}_\text{KV-col}(n)$.
When both inputs and features are not fully observed, we will see that both of these approaches result in biased estimates. 

\begin{figure*}[h]
    \centering
    \includegraphics[width=\textwidth]{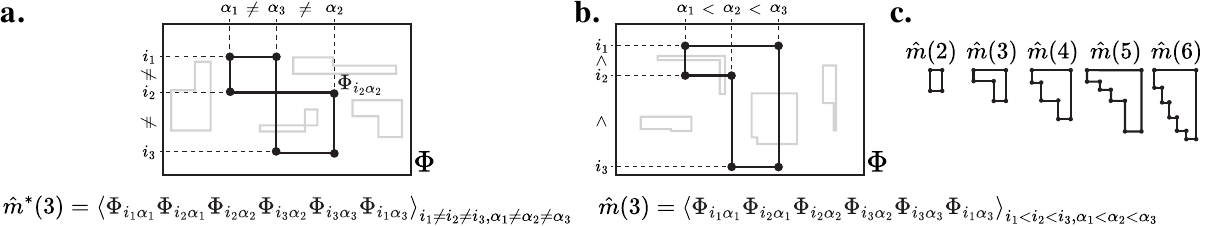}
    \caption{Visual illustration of the calculation of the unbiased estimator. \textbf{a.} For computing $\hat{m}^*(3)$, one can select matrix entries such that the entries create a cyclic path of 6 turns without revisiting rows and columns more than twice, and average over all possible such paths. \textbf{b.} Our method limits the cyclic paths to only increasing indices. \textbf{c.} Example paths for $\hat{m}(2),\ldots,\hat{m}(6)$.}
    \label{fig:diagram}
\end{figure*}

\section{UNBIASED ESTIMATION OF SPECTRAL MOMENTS AND EIGENVALUE RECOVERY}

\subsection{Unbiased estimator}
In the naive estimation of the second moment, $\hat{m}_0(2)$, we saw that certain terms in the sum were biased.  On the other hand, the first term of \eqref{eq:naive2} consisting of products of matrix coefficients with disjoint indices is unbiased. This observation can be generalized to higher spectral moments of the kernel integral operator. An elementary unbiased estimator for $m(n)$ is given by
\begin{equation}
\hat{m}'(n)=\prod_{l=1}^{n}\Phi_{i_{l}\alpha_{l}}\Phi_{i_{l+1}\alpha_{l}}
\end{equation}
where the indices $i_{l}\in\left\{ 1,\ldots,P\right\}$ are disjoint, $i_{l}\neq i_{k}$
for all $l\neq k$, except for the trace constraint $i_{1}=i_{n+1}$. Similarly, the feature indices
$\alpha_{l}\in\left\{ 1,\ldots,Q\right\} $ should also be disjoint, $\alpha_{l}\neq \alpha_{k}$ for
all $l\neq k$. Since there is no overlap in the indices, the expected value
of $\hat{m}'(n)$ is the product of the expected values of the
kernel functions:
\begin{align}
    \left\langle \prod_{l=1}^{n}\Phi_{i_{l}\alpha_{l}}\Phi_{i_{l+1}\alpha_{l}} \right\rangle _{\Phi} &= \prod_{l=1}^{n} \left\langle \Phi_{i_{l}\alpha_{l}}\Phi_{i_{l+1}\alpha_{l}} \right\rangle _{\Phi} \\
    &=\prod_{i=1}^{n}\left\langle k(x_{i},x_{i+1})\right\rangle _{x_{i},x_{i+1}} .
\end{align}
This is equivalent to the definition of $m(n)$, showing that $\hat{m}'(n)$ is
an unbiased estimator of $m(n)$.

As shown in Figure \ref{fig:diagram}, one can view the product in $\hat{m}'(n)$ as forming a cyclic path
over the coefficients of matrix $\Phi$ by first choosing a starting
coefficient $\Phi_{i_{1},\alpha_{1}}$, multiplying it with a distinct coefficient
$\Phi_{i_{2},\alpha_{1}}$ on the same column, then multiplying with another
coefficient $\Phi_{i_{2},\alpha_{2}}$ on the same row as the previous one,
and so on until returning to the starting coefficient, creating a product with a total of
$2n$ distinct coefficients.

In order to reduce the variance of this estimate, we can consider averaging $\prod_{l=1}^{n}\Phi_{i_{l}\alpha_{l}}\Phi_{i_{l+1}\alpha_{l}}$ over all
possible cyclic paths $\left\{ i_{l}\right\} _{l=1}^{n}$ and $\left\{ \alpha_{l}\right\} _{l=1}^{n}$ of non-overlapping indices:
\begin{equation}
\hat{m}^*(n)=\frac{
\sum_{\substack{\alpha_{1}\neq\ldots\neq \alpha_{n}\\ i_{1}\neq\ldots\neq i_{n}}} \prod_{l=1}^{n}\Phi_{i_{l}\alpha_{l}}\Phi_{i_{l+1}\alpha_{l}}}{\prod_{i=0}^{n-1}(P-i)(Q-i)},    
\end{equation}
with the constraint that the final input index is the same as the initial input index $i_{n+1}=i_1$.

This sum can be performed for small moments $n$.  For example, with $n=2$, we have the straightforward expression:
\begin{equation}
    c\hat{m}^*(2)=\hat{m}_0(2) -\sum_{i=1}^P\frac{K^2_{ii}}{P^2}
    -\sum_{\alpha=1}^Q\frac{\tilde{K}^2_{\alpha\alpha}}{Q^2}+\sum_{i=1}^P\sum_{\alpha=1}^Q\frac{\Phi^4_{i\alpha}}{P^2Q^2}
\end{equation}
where $c\coloneqq\frac{(P-1)(Q-1)}{PQ}$, $K_{ij}=\frac{1}{Q}\sum_{\alpha=1}^{Q}\Phi_{i\alpha}\Phi_{j\alpha}$ and $\tilde{K}_{\alpha \beta}=\frac{1}{P}\sum_{i=1}^{P}\Phi_{i\alpha}\Phi_{i\beta}$.

Unfortunately, summing over all disjoint index sets is computationally
inefficient for larger $n$, with apparent complexity $\mathcal{O}(P^{n}Q^{n})$. Instead,
similar to the approach by \cite{kong2017}, we can average over cyclic paths
where both sets of indices are increasing, i.e. $1\le i_{1}<i_{2}<\ldots<i_{n}\le P$
and $1\le \alpha_{1}<\alpha_{2}<\ldots<\alpha_{n}\le Q$ (see Figure \ref{fig:diagram}b):

\begin{equation}\label{eq:incm}
\hat{m}(n)=\frac{1}{\binom{P}{n}\binom{Q}{n}}
\sum_{\substack{1\le i_{1}<i_{2}<\ldots<i_{n}\le P \\ 1\le \alpha_{1}<\alpha_{2}<\ldots<\alpha_{n}\le Q}}
\prod_{l=1}^{n}\Phi_{i_{l}\alpha_{l}}\Phi_{i_{l+1}\alpha_{l}}
\end{equation}
where $i_{n+1}=i_1$,
and the combinatorial product $\binom{P}{n}\binom{Q}{n}$ is the total
number of terms in the sum. Graphically, the paths over increasing index sets create stair-like cyclic paths in the matrix $\Phi$ as shown in Figure \ref{fig:diagram}c.

\subsection{Dynamic programming algorithm}

\begin{algorithm}[t!]
\caption{Computation of $\hat{m}(n)$ for $n = 2$ to $n_{\text{max}}$}
\begin{algorithmic}[1]
\Require $\Phi \in \mathbb{R}^{P \times Q}$, $n_{\text{max}}$
\For{$h \leftarrow 1$ \textbf{to} $P$}
    \State Initialize $S$ as a $P \times Q$ zero matrix.
    \State Set $S_{hi} \leftarrow P \Phi_{hi}$  $\forall i \in [1, Q]$
    \For{$n \leftarrow 2$ \textbf{to} $n_{\text{max}}$}

        \State \begin{varwidth}[t]{\columnwidth} Update \par
        $S_{ab} \leftarrow 
        \dfrac{n^{2}\sum_{l=h+n-2}^{a-1} \sum_{k=n-1}^{b-1}  S_{lk} \Phi_{ak} \Phi_{ab}}{(P - n + 1)(Q - n + 1)}$ \par $\forall a \in [h+n-1, P]$, $\forall b \in [n, Q]$.
        \end{varwidth}
        \label{algo:update}
        
        \State \begin{varwidth}[t]{\columnwidth}
        Compute \par $\hat{m}^{(h)}(n) \leftarrow \dfrac{1}{PQ} \sum_{i=h+n-1}^{P} \sum_{j=n}^{Q} S_{ij} \Phi_{hj}$.
        \end{varwidth}
    \EndFor
\EndFor
\State Get $\hat{m}(n) \gets \dfrac{1}{P} \displaystyle\sum_{h=1}^{P-n+1} \hat{m}^{(h)}(n)$ $\forall n \in [2,n_\text{max}]$.
\end{algorithmic}
\label{algo}
\end{algorithm}

Here we show how to efficiently compute 
$\hat{m}(n)$ for higher moments via a recursive dynamic programming algorithm. 
The estimate $\hat{m}(n)$ can be written in terms of the partial sums:
\begin{equation}\label{eq:dpm}
\hat{m}(n)=\frac{1}{\binom{P}{n}\binom{Q}{n}}\sum_{h=1}^{P-n+1}\sum_{j=n}^{Q}\sum_{i=h+n-1}^{P}S_{ij}^{(h)}[n]\Phi_{hj}.
\end{equation}
where the sums are defined as:
\begin{multline}
S_{ab}^{(h)}[n]=\sum_{\substack{h<i_{2}<\ldots<i_{n-1}<a \\ 1\le \alpha_{1}<\ldots<\alpha_{n-1}<b}}
\left(\prod_{l=1}^{n-1}\Phi_{i_{l}\alpha_{l}}\Phi_{i_{l+1}\alpha_{l}}\right)\Phi_{i_{n}\alpha_{n}}
\end{multline}
The indices $h$, $a$, $b$ in $S_{ab}^{(h)}$ correspond to the indices $i_1$, $i_n$, and $\alpha_n$ in the estimator expression \eqref{eq:incm}.
Next, we note that $S_{ab}^{(h)}$ can be written recursively:
\begin{equation}\label{eq:dp}
S_{ab}^{(h)}[n+1]=\sum_{k=n}^{b-1}\sum_{l=h+n-1}^{a-1}S_{lk}^{(h)}[n]\Phi_{ak}\Phi_{ab}.
\end{equation}
In this manner, we can iteratively compute $S^{(h)}[n]\to S^{(h)}[n+1]$ to obtain all the partial sums for the spectral moment estimation.

This computation is also memory efficient. The partial sums 
$S_{lk}^{(h)}[n]$ can be stored as the $(lk)$-th element of a $P\times Q$ matrix $\left[ S^{(h)}_{lk}[n] \right]$ and the computation can be performed in place. The algorithm first initializes the matrix $S^{(h)}[1]$ by setting its $h$-th row to match the $h$-th row of $\Phi$, with the rest of the elements initialized to $0$. Then $S^{(h)}[2]$ is computed via \eqref{eq:dp} to obtain $\hat{m}(2)$ with \eqref{eq:dpm}. This procedure can be repeated to get $\hat{m}(n)$ for all $n$ ranging from $2$ to the desired $n_\text{max}$. The estimate for $m(1)$ is the same as the naive estimate $\hat{m}_0(1)$ which is unbiased. Pseudo-code of our recursive algorithm to compute the spectral moment estimates is provided in Algorithm \ref{algo}.

The computational complexity of our algorithm is $\mathcal{O}(nP^{2}Q)$, or $\mathcal{O}(nPQ^2)$ if the algorithm is performed on the matrix transpose $\Phi^\top$. In practice, $S^{(h)}[n]$ should also be normalized at each of step of the recursion to prevent any overflow in the calculation; this normalization is included in the description of Algorithm \ref{algo}. Additionally, step \ref{algo:update} in Algorithm \ref{algo} can readily be vectorized with a simple modification to the cumulative summation subroutine.  To further improve the accuracy of the estimates, the rows and columns of $\Phi$ can first be permuted and the same algorithm can be performed to compute additional cyclic path sums to reduce the variance of the spectral moment estimates.

\subsection{Noisy measurements}

Our estimator $\hat{m}(n)$ is unbiased even in the presence of independent noise, when the noise is injected into the generative process as $\Phi_{i\alpha} = \phi(x_i,w_\alpha)+\gamma(x_i,w_\alpha,\epsilon_{i\alpha})$, assuming: $\epsilon_{i\alpha}$ is sampled independently from some probability measure $\rho_\epsilon$ across all $(i,\alpha)$; $\langle\gamma(x,w,\epsilon)\rangle_\epsilon=0$; and $\langle\gamma(x,w,\epsilon)^2\rangle_\epsilon<\infty$.

We can also handle the case when the noise $\epsilon_{i\alpha}$ is correlated across row or column entries of $\left[\Phi_{i\alpha}\right]$ with a simple modification to our estimator. Multiple measurements can be taken over the same set of inputs and features, $\left\{x_i \right\}_{i=1}^P$ and $\left\{w_\alpha\right\}_{\alpha=1}^Q$, and can be used to form separate trial measurements of the matrix $\Phi^{(t)}$ with $t$ indexing different trials with independent noise across trials. With measurement samples from only two trials $\Phi^{(1)}$ and $\Phi^{(2)}$, unbiased estimates of the moments can be obtained by alternating trial measurements in the product terms:
$\hat{m}'_\text{alt}(n)=\prod_{l=1}^{n}\Phi^{(1)}_{i_{l}\alpha_{l}}\Phi^{(2)}_{i_{l+1}\alpha_{l}}
$. This procedure can be easily extended with additional trials to further denoise the spectral moment estimates, and the corresponding modifications to our dynamic programming algorithm are straightforward (see Appendix for more details).

For the rest of the theoretical and experimental analyses in the main text, we assume the measurement is noise-free.

\subsection{Variance of $\hat{m}(n)$}
We now bound the variance of our estimator $\hat{m}(n)$ and derive a probabilistic guarantee for its accuracy. The following lemma gives an upper bound when $\hat{m}(n)$ is computed from a $P\times Q$ measurement matrix.

\begin{theorem}\label{varthe}
Suppose $\phi\in\mathcal{L}^4(\rho_\mathcal{X}\otimes \rho_\mathcal{W})$. Then variance of $\hat{m}(n)$ satisfies
\begin{equation}
\operatorname{Var}\!\left(\hat{m}(n)\right) \le \left(\frac{1}{P} + \frac{1}{Q}\right) f(n),    
\end{equation}
where
\begin{equation}
f(n) = n^{2}\operatorname{Var}\!\Biggl(\prod_{i=1}^{n}\phi(x_{i},w_{i})\,\phi(x_{i+1},w_{i})\Biggr).    
\end{equation}

\end{theorem}

Applying Chebyshev's inequality yields the following guarantee on the absolute error:
\begin{corollary}\label{corr1}
Suppose $\phi\in\mathcal{L}^4(\rho_\mathcal{X}\otimes \rho_\mathcal{W})$. Then for any $\delta \in (0,1)$, with probability at least $1-\delta$ we have
\begin{equation}
\bigl|\hat{m}(n)-m(n)\bigr| \le \sqrt{\frac{f(n)}{\delta}\left(\frac{1}{P}+\frac{1}{Q}\right)}.    
\end{equation}
\end{corollary}
Our estimator is also strongly consistent. The complete proofs of Theorem~\ref{varthe} and estimator consistency are provided in the Appendix.


\subsection{Eigenvalues from Moments}
\cite{kong2017} presents a linear programming method to recover a finite sequence of eigenvalues from estimated spectral moments. The algorithm takes as input the moments $\{\hat{m}(n)\}_{n=1}^k$, the number $d$ of eigenvalues to recover, and an upper bound $b$ on the eigenvalues. It approximates the spectral density $p$ with a discrete distribution $\hat{p}$ defined on points
$\mathcal{S} \coloneqq \{s_i\}_{i=1}^T \subset [0,b]$,
and then returns the $(d+1)$st quantiles of $\hat{p}$ as the eigenvalue estimates.

To compute $\hat{p}$, the algorithm minimizes
\begin{equation}
\min_{\{\hat{p}_i\}_{i=1}^T} \sum_{n=1}^k \Biggl|\hat{m}(n) - \sum_{i=1}^T \hat{p}_i\, s_i^n \Biggr|,    
\end{equation}
subject to $\sum_{i=1}^T \hat{p}_i = 1$ and $\hat{p}_i \ge 0\quad \forall\, i$.
This optimization is solved via linear programming.

Assuming the kernel integral operator has finite rank $d$, we obtain the following error bound for the recovered eigenvalues.

\begin{corollary}
Suppose the kernel integral operator has rank $d<\infty$. Then, the expected total absolute error in the recovered eigenvalues $\{\lambda_i\}_{i=1}^d$ via the method in \cite{kong2017} applied to our moments $\{\hat{m}(n)\}_{n=1}^k$ is bounded by
\begin{multline}
\left\langle \sum_{i=1}^{d}\bigl|\lambda_{i} - \hat{\lambda}_{i}\bigr| \right\rangle \le \\
bd\left(c\, 3^{k}k^{2}\left(\sqrt{\left(\frac{1}{P}+\frac{1}{Q}\right)\frac{f(k)}{k^{2}}} + dc'\right) + \frac{c''}{k} + \frac{1}{d}\right),
\end{multline}
where $c$, $c'$, and $c''$ are positive constants.
\end{corollary}

The proof is provided in the Appendix.

\section{RBF KERNEL OPERATOR}

In this section, we describe the spectrum of the kernel integral operator for the radial basis function (RBF) kernel function with multivariate Gaussian inputs. We then confirm that our method estimates the correct spectral moments and compare its performance to other methods.

\subsection{Random Fourier features}
\citet{rahimi2007random} and \citet{rudi2017generalization} describe the following process using random Fourier features. Consider a two-layer neural network with an input layer and a feature layer with a sinusoidal nonlinearity. Given an input pattern $x\in\mathbb{R}^{d}$ in the input layer, the value of a random feature in the feature layer is defined by weights $w\in\mathbb{R}^{d}$ and phase shift $b\in\mathbb{R}$ as:
\begin{equation}\label{eq:RFF}
\phi(x,(w,b))=\sqrt{2}\sin\left(w^{\top}x+b\right)
\end{equation}
\begin{equation}
w\sim\mathcal{N}\left(0,\Sigma^{-1}\right)
\qquad
b\sim\mathcal{U}\left(0,2\pi \right)
\end{equation}
so that $w$ is a random weight vector sampled from a multivariate normal
distribution $\mathcal{N}\left(0,\Sigma^{-1}\right)$, and $b$ is
a random phase shift sampled from the uniform distribution ${\mathcal U}$.
The similarity between two inputs $x$ and $y$ is given by taking the expectation over all possible features and is
equivalent to the RBF kernel:
\begin{equation}
k(x,y)=e^{-\frac{1}{2}\left(x-y\right)^{\top}\Sigma^{-1}\left(x-y\right)}.
\end{equation}
In the following, we also assume that the inputs $x$ and $y$ are sampled from an input distribution $\rho_\mathcal{X}(x)$ described by the multivariate 
normal distribution ${\mathcal N}(0,\Sigma_{x})$.

\subsection{Spectrum of the RBF kernel}

\begin{figure*}[h]
    \begin{center}
        \resizebox{1\linewidth}{!}{\input{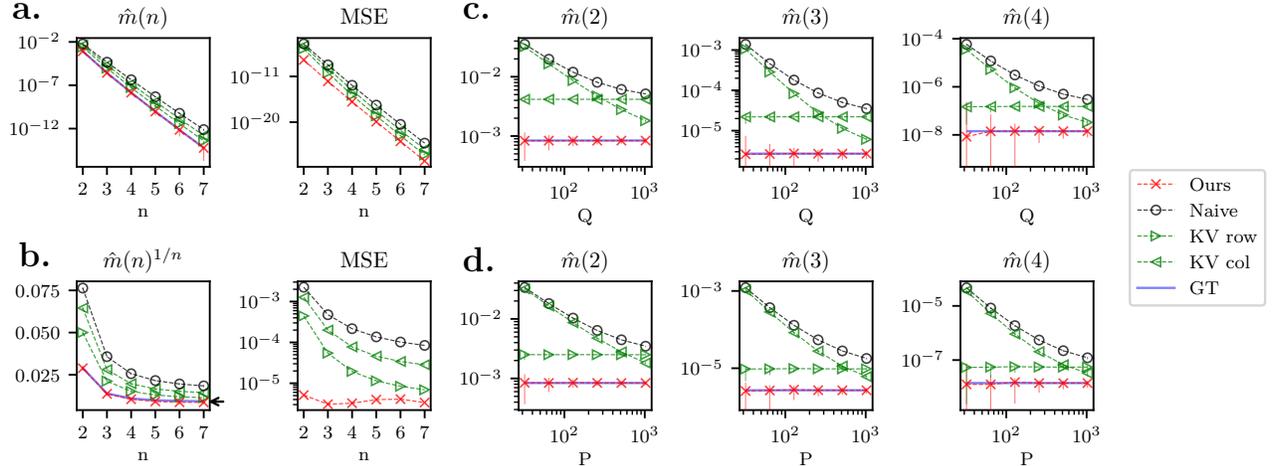}}
    \end{center}
    \caption{ Estimated RBF moments for $d=5$, $\Sigma_x=I_{d\times d}$, $\Sigma=0.25I_{d\times d}$. Our estimator $\hat{m}$ is labeled as ``Ours", the two versions of Kong and Valiant estimators $\hat{m}_{\text{KV-row}}$ and $\hat{m}_{\text{KV-col}}$ are labeled as ``KV-row" and ``KV-col" respectively, the naive estimator $\hat{m}_0$ is labeled as ``naive", and the analytic ground truth moments $m$ are labeled as ``GT". \textbf{a.} $P=300$ and $Q=600$. Left: The $\hat{m}(n)$ values for various estimators, with $n$ ranging from 2 to 7. Right: The MSE between $\hat{m}(n)$ and $m(n)$. \textbf{b.} The same as \textbf{a.}, but for $\hat{m}^{1/n}(n)$. The black arrow indicates the value of the operator norm. \textbf{c.} $P$ is fixed to 300, and $Q$ is varied. \textbf{d.} $Q$ is fixed to 600, and $P$ is varied. Bars indicate a $50\%$ confidence interval.}
    \label{fig:RBF}
\end{figure*}

The spectrum of the RBF kernel operator has been described before for $d=1$, or when the kernel and input covariances are isotropic, or can be simultaneously factorized in \citet{zhu1997gaussian,williams2000effect,williams2006gaussian,canatar2021spectral}. 
The spectrum of the general kernel integral operator for arbitrary kernel and input covariances is described in terms of the $d\times d$ positive-definite matrix $\Sigma_{x}\Sigma^{-1}$. Let $\left\{ \eta_{i}\right\} _{i=1}^{d}$ be the eigenvalues of $\Sigma_{x}\Sigma^{-1}$ and let $u\coloneqq\left\{ u_{i}\right\} _{i=1}^{d}$
be a multiset of $d$ natural numbers. Then the following is an eigenvalue
of the kernel integral operator for all $u$:
\begin{equation}
\lambda_{u}=\prod_{i=1}^{d}\left(\eta_{i}^{1+u_{i}}
\varphi_{\eta_i}^{1+2u_i}
\right)^{-1}
\end{equation}
where the scalar function $\varphi_z = \frac{1+\sqrt{1+4z}}{2z}$.
The largest kernel operator eigenvalue, e.g. the spectral
norm of the operator, is obtained when $u_{i}=0$ for all $i$. The corresponding eigenfunctions are given by products of generalized Hermite polynomials with degree $u_i$ and a common multivariate Gaussian function (see Appendix).

The $n$-th spectral moments can then be computed as
\begin{equation}\label{eq:RBF_moments}
m(n)=\prod_{i=1}^{d}\frac{1}{\eta_{i}^{n}\varphi_{\eta_{i}}^{n}-\varphi_{\eta_{i}}^{-n}} .
\end{equation}
Note that $m(1)=1$ regardless of the choice of kernel parameters.

A particularly interesting case is when $\Sigma=\Sigma_{x}$ so that 
$\eta_{i}=1$ for all $i$. In this case, the kernel operator eigenvalues are powers
of the golden ratio $\varphi_{1}$, i.e. $\lambda_{t}=\varphi_{1}^{-t}$
for $t\in\{1,2,\ldots\}$, where the $t$-th eigenvalue has multiplicity
$\binom{t+d-1}{t}$. The moments are given by $m(n)=\left(\frac{1}{\phi_{1}^{n}-\phi_{1}^{-n}}\right)^{d}$, and decrease exponentially for $n\geq 2$ to zero for larger $d$.

\subsection{Numerical estimation results}
We consider $P\times Q$ measurement matrices $\left[\Phi_{i\alpha}\right]$ taking $P$ samples of inputs patterns $\left\{ x_i \right\}_{i=1}^P$ and $Q$ samples of weights and phase shifts $\left\{(w_\alpha,b_\alpha )\right\}_{\alpha=1}^Q$. The naive estimator $\hat{m}_0(n)$, Kong and Valiant's estimators $\hat{m}_\text{KV-row}(n)$ and $\hat{m}_\text{KV-col}(n)$, and our method, $\hat{m}(n)$, are all applied to estimate the spectral moments of the operator and compared with the analytical formula in \eqref{eq:RBF_moments}. As shown in Figure \ref{fig:RBF}a, there is excellent agreement between our estimates (red dotted lines) and the true moments (blue solid lines). Our estimator achieves the smallest mean-squared error, showing that it is both unbiased and has low variance (see Appendix for detailed bias-variance analysis of the various estimators along with their performance in the presence of independent and correlated noise).

The difference between the estimators is more pronounced when comparing the $n$-th root of the estimated $n$-th moment, i.e. $m(n)^{1/n}$. $m(2)^{1/2}$ is the standard deviation of the spectrum, and $\lim_{n\to\infty} m(n)^{1/n}$ gives the operator norm, the largest eigenvalue of the kernel integral operator. Note that even though $\hat{m}(n)$ is unbiased, the $n$-th root of the estimate, $\hat{m}(n)^{1/n}$, will be a biased estimate of $m(n)^{1/n}$. Nevertheless, we observe that our estimates of the rooted moment achieve remarkably small MSE ($<10^{-5}$, Figure \ref{fig:RBF}b right). We also observe that our rooted moment estimate accurately recovers the operator norm (Figure \ref{fig:RBF}b left, compare $\hat{m}(7)$ with the black arrow).

Our estimator can be compared to the other estimators as both $P$ and $Q$ are varied. If we fix $P$ and vary $Q$, $\hat{m}_0(n)$ and $\hat{m}_\text{KV-col}(n)$ are biased even in the limit of large $Q$, due to the small sampling of $P$ (Figure \ref{fig:RBF}c). On the other hand, our estimator and $\hat{m}_\text{KV-col}(n)$ asymptotically converge to the true moments at large $Q$. Similarly,  varying $P$ for fixed $Q$ (Figure \ref{fig:RBF}) shows that only our estimator and $\hat{m}_\text{KV-row}(n)$ asymptotically converges to the true moments. Our estimator is the only unbiased estimator across all finite $P$ and $Q$.

\section{EIGENVALUE RECONSTRUCTION}

\begin{figure}
    \begin{minipage}[t]{0.49\linewidth}
        \centering
        \scalebox{0.8}{\input{./figures/lin_eigvals.pgf}}
    \end{minipage}
    \begin{minipage}[t]{0.49\linewidth}
        \centering
        \scalebox{0.8}{\input{./figures/RBF_eigvals.pgf}}
    \end{minipage}
    \caption{The reconstructed eigenvalues of two generative processes. ``GT" refers to ground truth eigenvalues. ``SVD" is the singular values of the empirical Gram matrix. ``KV" is the eigenvalues reconstructed from $\left\{\hat{m}_\text{KV}(n)\right\}_{n=1}^{10}$. ``Ours" is the eigenvalues reconstructed from $\left\{\hat{m}(n)\right\}_{n=1}^{10}$.  Left: finite-rank linear generative process whose true eigenvalue is 0.3 with multiplicity 20. $P=Q=100$. Right: Random Fourier feature generative process whose true $i$th eigenvalue is $\left(\eta \varphi_\eta\right)^{-i}$ with $\eta=400$. $P=Q=20$.} 
    \label{fig:eigen}
\end{figure}

Here we demonstrate the recovery of the eigenvalues of kernel integral operators from finite samples by combining our moment estimator $\hat{m}$ with the moment-to-spectrum algorithm of \cite{kong2017}.

In the first example, consider the generative process
\begin{equation}
\phi(x,w) = \sqrt{0.3}\, x^\top w,\quad \rho_\mathcal{X} = \rho_\mathcal{W} = \mathcal{N}(0,I_{d\times d}).
\end{equation}
The corresponding kernel operator has finite rank $d$ with all nonzero eigenvalues equal to $0.3$. As shown in Figure~\ref{fig:eigen} (Left), our estimator $\hat{m}$ yields an eigenvalue spectrum that closely matches the true spectrum, while the reconstruction based on $\hat{m}_\text{KV}$ is significantly biased.

In the second example, we recover the eigenvalues of the RBF kernel derived from finite random Fourier features. Here, the kernel operator has a countably infinite number of exponentially decaying eigenvalues. Figure~\ref{fig:eigen} (Right) shows that the largest $d=50$ eigenvalues are accurately recovered using our estimator $\hat{m}$, whereas other methods fail. Note that in this case, the estimation is sensitive to the choice of parameters $d$ and $T$.

These examples confirm that our moment-based approach effectively reconstructs the eigenvalue spectrum of kernel integral operators from finite measurements.

\section{RELU KERNEL MOMENTS DURING FEATURE LEARNING}

\begin{figure*}[h]
    \begin{center}
        \resizebox{1\linewidth}{!}{\input{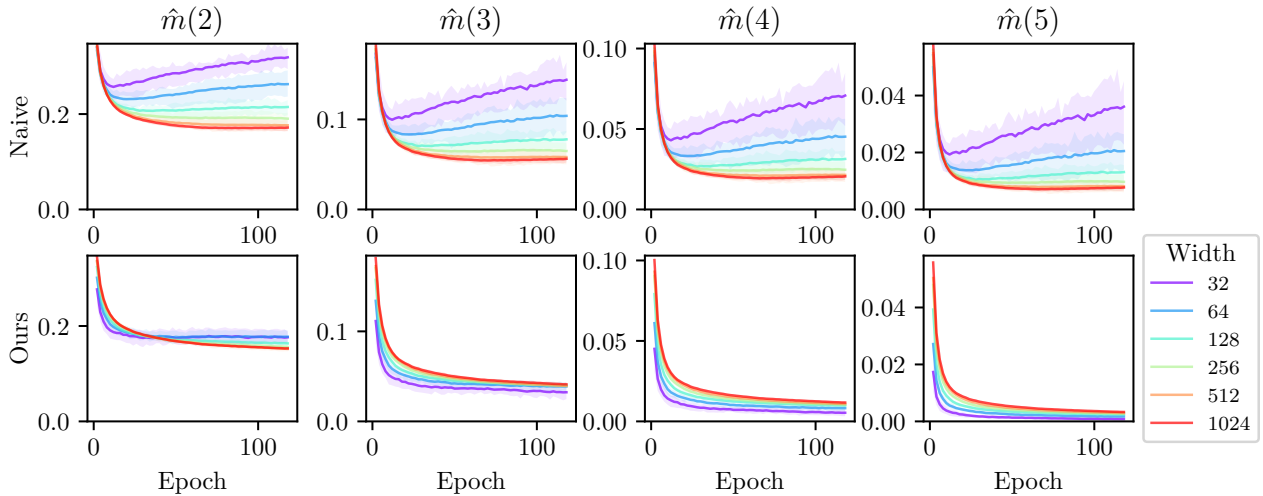}}
    \end{center}
    \caption{The estimated spectral moments during training of single hidden layer ReLU neural networks. Top row: networks of different widths have dramatically different naive estimates $\hat{m}_0(n)$ of the operator moments. Bottom row: estimates using the unbiased estimator $\hat{m}(n)$ is similar across all widths. Results were obtained from networks trained from 29 random initializations. Shades indicate a $50\%$ confidence interval.}
    \label{fig:trained}
\end{figure*}

Here we show how our estimator can be used to analyze the neural representation of varying widths in a neural network during feature learning. Specifically, a single-hidden layer neural network with ReLU activation, trained on the Fashion-MNIST dataset with Adam-SGD \citep{xiao2017Fashion,kingma2014adam} is considered. To enable efficient feature learning, the maximal-update parameterization ($\mu$P) as proposed by \citet{yang2022tensor} is utilized. As the width becomes large, the trained network approaches the mean-field limit where each feature becomes an i.i.d. random variable in the learned weight distribution, and consequently remains exchangeable after training \citep{mei2018mean,yang2022tensor,vyas2024feature,bordelon2022self,seroussi2023separation,yang2023theory}. 

The measurement matrix $\Phi$ is constructed from a network with $Q$ neurons in the hidden layer, and each row of $\Phi$ represents the $Q$-dimensional hidden layer activations from one of $P$ input images. $\Phi$ is normalized so that it can be compared across different matrix sizes by dividing by the standard deviation of its entries.
The spectral moments of the hidden layer representations for networks with widths ranging from $32$ to $1024$ are shown in Figure \ref{fig:trained}. The naive spectral moment estimates diverge across the different network widths; in contrast, our estimator produces consistent spectral moments across the entire range of widths. This indicates that predictions of learning dynamics from feature learning theories can be applied to understand the behavior of neural networks across a wide range of sizes. We defer exploration of how feature learning theories such as mean-field theory can be used to model our spectral moment estimates to future work.

\section{DISCUSSION}

We have shown that conventional methods for analyzing the spectrum of a measurement matrix with finitely sampled inputs and features are biased. As an alternative, we propose an unbiased method for estimating the spectral moments of the kernel integral operator from finite measurement matrices. Our method is computationally efficient and results in accurate moment and eigenvalue estimates, as demonstrated in numerical experiments with the RBF kernel where analytic results for the true operator spectrum are known.

Our estimator can be used to gain geometrical insight into measurement matrices of varying sizes. For example, we can accurately estimate the effective dimension of the kernel operator $T_k$. By considering $\text{tr}\left[ T_k\left(T_k +\lambda I\right)^{-1}\right]$ whose Taylor series include weighted sums of the spectral moments \citep{caponnetto2007optimal,bach2013sharp}, a soft count of the number of eigenvalues above a threshold $\lambda$ can be obtained. This quantity has been widely used to study prediction performance in kernel ridge regression.

The spectral moment estimates can also be employed in conjunction with methods to approximate kernels with sampled features to reduce computational complexity \citep{rahimi2007random,rudi2017generalization,rudi2024finding}. For example, kernel approximation has recently been used in state-of-the-art transformer models by replacing the softmax attention with a kernel \citep{peng2021random}. 
The decay of the kernel integral operator spectrum can be used to gauge the accuracy of the kernel approximation. 

We showed how our estimator can be used to quantify the learning dynamics of neural networks during training. The kernel operator can be related to the Hessian of a quadratic objective function \citep{dieuleveut2017harder,pedregosa2020acceleration,sagun2017empirical,martin2021implicit} which is valuable for understanding gradient-based learning dynamics. Other recent work proposes spectral estimates to understand convergence rates \citep{dieuleveut2017harder} as well as accelerating the optimization \citep{pedregosa2020acceleration}.
Thus, there are many potential avenues for future exploration where unbiased and efficient estimates of the spectral characteristics of the kernel integral operator will be valuable.

\subsubsection*{Acknowledgements}
We thank Jonathan D. Victor and Abdulkadir Canatar for their valuable feedback on the project. This work was supported by the Simons Foundation.

\bibliography{biblio}

\clearpage
\appendix

\renewcommand{\thefigure}{A\arabic{figure}}
\setcounter{figure}{0}

\renewcommand{\thealgorithm}{A\arabic{algorithm}}
\setcounter{algorithm}{0}

\renewcommand{\theequation}{A\arabic{equation}}
\setcounter{equation}{0}

\onecolumn

\section{Kernel integral operator moments and Stieltjes transform}
We can easily see that the sequence of spectral moments $\left\{ \sum_{i=1}^\infty \lambda_i^n \right\}_{n=1}^\infty$ uniquely determines the non-zero eigenvalues $\left\{ \lambda_i \right\}_{i=1}^\infty$ of a self-adjoint trace-class operator.
Consider the following Stieltjes transform where $z$ is in the complex plane:
\begin{equation}
g(z)=\sum_{i=1}^{\infty}\frac{\lambda_{i}}{z-\lambda_{i}}    
\end{equation}
Its Taylor series for $z^{-1}$ near zero is

\begin{equation}
    g(z)=\sum_{i=1}^\infty \lambda_{i}z^{-1} + {\lambda_i^2}z^{-2} +{\lambda_i^3}z^{-3} + {\lambda_i^4}z^{-4} + \mathcal{O}\left(z^{-5}\right)
\end{equation}
which is equivalent to 
\begin{equation}
    g(z)= m(1) z^{-1} + m(2)z^{-2} +m(3)z^{-3} +m(4)z^{-4} +\ldots
\end{equation}
with moments $m(n)=\sum_{i=1}^\infty \lambda_i^n $. The moments $\left\{ m(n) \right\}_{n=1}^\infty$ uniquely define the complex meromorphic function $g(z)$, and the eigenvalues  $\left\{\lambda_i\right\}_{i=1}^\infty$ can then be determined by the location of the poles of $g(z)$. Therefore, the operator moments uniquely determine the non-zero operator eigenvalues.

Newton's identities can also be used to express the relationship between operator moments and the characteristic equation for the eigenvalues of finite rank $d$ operators.
Consider the characteristic polynomial
\begin{equation}
f(\lambda)=\prod_{i=1}^d \left(\lambda-\lambda_i \right)    
\end{equation}
with roots at the eigenvalues of the operator, i.e. $\{\lambda_i\}_{i=1}^d$. The function $f(\lambda)$ can be decomposed in decreasing orders of $\lambda$ with coefficients consisting of the elementary symmetric polynomials of the roots.  Newton's identities recursively relate these coefficients with the power sums of $\lambda_i$, which are identical to the spectral moments:  $\{m(n)\}_{n=1}^d$. Thus, the spectral moments uniquely determine the characteristic polynomial of the eigenvalues.

\section{Kernel covariance operator}

Let $\rho_{{\mathcal X}}(x)$ and $\rho_{{\mathcal W}}(w)$ be probability
measures over latent spaces $\mathcal{X}$ and $\mathcal{W}$ respectively.
The map $\phi:{\mathcal X}\times\mathcal{W}\to\mathbb{R}$ is square-integrable
with respect to both $\rho_{{\mathcal X}}$ and $\rho_{{\mathcal W}}$ and
determines the $\left(i\alpha\right)$-th coefficient of the measurement
matrix $\left[\Phi_{i\alpha}\right]$ by $\Phi_{i\alpha}=\phi(x_{i},w_{\alpha})$.
Let $\psi(x)\coloneqq\phi(x,\cdot):\mathcal{W}\to\mathbb{R}$. Then
the completion of the linear space of the functions $\left\{ \psi(x)\right\} _{x\in{\mathcal X}}$
is $\mathcal{F}\subset \mathcal{L}^{2}(\mathcal{W},\rho_{\mathcal{W}})$, with
the inner product 
\begin{equation}
\left\langle f\vert f^{\prime}\right\rangle _{\mathcal{F}}=\int d\rho_{\mathcal{W}}(w)\:f(w)f^{\prime}(w)
\end{equation}
where $f,f'\in\mathcal{F}$. Now consider the covariance operator
$T_{c}:\mathcal{F}\rightarrow\mathcal{F}$,
\begin{equation}
T_{c}\coloneqq\int d\rho_{\mathcal{X}}(x)\,\left|\psi(x)\right\rangle \left\langle \psi(x)\right|
\end{equation}
$\mathcal{F}$ is also a reproducing kernel Hilbert space (RKHS),
so the Riesz representation theorem implies that for all $w\in\mathcal{W},$
there exists a unique evaluation function $\varphi(w)\in\mathcal{F}$
such that $f(w)=\left\langle f\vert\varphi(w)\right\rangle _{\mathcal{F}},\forall f\in\mathcal{F}$.
The map $\phi$ can then be expressed as an inner product in the RKHS, $\phi(x,w)\equiv\left\langle \psi(x)\vert\varphi(w)\right\rangle _{\mathcal{F}}$.
Define the frame operator $S:\mathcal{F}\rightarrow\mathcal{F}$:
\begin{equation}
S=\int d\rho_{\mathcal{W}}(w)\:\left|\varphi(w)\right\rangle \left\langle \varphi(w)\right|
\end{equation}
Then $S$ is equivalent to the identity operator, since $\left\langle f\vert S\vert f^{\prime}\right\rangle _{\mathcal{F}}=\left\langle f\vert f^{\prime}\right\rangle _{\mathcal{F}}$.
The set of $\left|\varphi(w)\right\rangle $ is a tight Parseval frame
in $\mathcal{F}$.

Now we can interpret what it means to compose $T_{c}$:
\begin{equation}
T_{c}^{n}=\int\prod_{l=1}^{n}d\rho_{\mathcal{X}}(x_{l})\,\left|\psi(x_{1})\right\rangle \langle\psi(x_{1})\vert\psi(x_{2})\rangle\cdots\langle\psi(x_{n-1})\vert\psi(x_{n})\rangle\langle\psi(x_{n})\vert
\end{equation}
the trace of which is
\begin{equation}
\mathrm{tr}T_{c}^{n}=\int\prod_{l=1}^{n}d\rho_{\mathcal{X}}(x_{l})\,\prod_{l=1}^{n}\left\langle \psi(x_{l})\vert\psi(x_{l+1})\right\rangle 
\end{equation}
with the trace constraint $x_{n+1}=x_1$. Since $S$ is the identity operator
on $\mathcal{F}$, $\langle\psi(x)\vert\psi(x')\rangle$ is equivalent
to $\langle\psi(x)\vert S\vert\psi(x')\rangle$, so 
\begin{equation}
\langle\psi(x)\vert\psi(x')\rangle\equiv\int d\rho_{\mathcal{W}}(w)\:\left\langle \psi(x)\vert\varphi(w)\right\rangle _{\mathcal{F}}\left\langle \psi(x')\vert\varphi(w)\right\rangle _{\mathcal{F}}.
\end{equation}
Therefore, the trace becomes
\begin{equation}
\mathrm{tr}T_{c}^{n}=\int\prod_{l=1}^{n}d\rho_{\mathcal{X}}(x_{l})d\rho_{\mathcal{W}}(w_{l})\,\prod_{l=1}^{n}\phi(x_{l},w_{l})\phi(x_{l+1},w_{l}).
\end{equation}
Now we can relate $T_{c}$ to $T_{k}$ defined in the main text. With
the definition of the kernel function and $T_{k}:\mathcal{L}^{2}(\mathcal{X},\rho_{\mathcal{X}})\to \mathcal{L}^{2}(\mathcal{X},\rho_{\mathcal{X}})$,
we see that

\begin{equation}
    \mathrm{tr}T_{c}^{n}=
\int\prod_{i=1}^{n}d\rho_{\mathcal{X}}(x_{i})\,\prod_{i=1}^{n}k(x_{i},x_{i+1})
    \equiv
    \text{tr}T_{k}^{n}.
\end{equation}

\section{Proof of estimator consistency}

Consider the sets of increasing index sequences:
\begin{equation}
\{i_1, i_2, \dots, i_n\} \subset \{1, 2, \dots, P\}, \quad \text{where} \quad i_1 < i_2 < \dots < i_n
\end{equation}
and similarly for $\{\alpha_1, \alpha_2, \dots, \alpha_n\} \subset \{1, 2, \dots, Q\}$, where $\alpha_1 < \alpha_2 < \dots < \alpha_n$. There are $\binom{P}{n}$ such sequences for $\{i_l\}$ and $\binom{Q}{n}$ for $\{\alpha_l\}$.

Define
\begin{equation}
\hat{m}(n) = \frac{1}{\binom{P}{n} \binom{Q}{n}} \sum_{1 \leq i_1 < \dots < i_n \leq P} \sum_{1 \leq \alpha_1 < \dots < \alpha_n \leq Q} \prod_{l=1}^n \Phi_{i_l \alpha_l} \Phi_{i_{l+1} \alpha_l}
\end{equation}
where $i_{n+1} = i_1$. The estimator $\hat{m}(n)$ is an unbiased estimator of
\begin{equation}
    m(n)\coloneqq\int \prod_{l=1}^n d\rho_\mathcal{X}(x_l) d\rho_\mathcal{W}(w_l)  \prod_{l=1}^n \phi(x_l, w_l) \phi(x_{l+1}, w_l).
\end{equation}

\begin{theorem}
$\hat{m}(n)$ is a strongly consistent estimator of $m(n)$, namely
\begin{equation}
    \hat{m}(n) \xrightarrow{\text{a.s.}} m(n)
\end{equation}
as $P, Q \to \infty$.    
\end{theorem}

\begin{proof}
Define the function
\begin{equation}
h(\{w_\alpha\} \mid \{x_i\}) = \frac{1}{\binom{P}{n} \binom{Q}{n}} \sum_{1 \leq i_1 < \dots < i_n \leq P} \sum_{1 \leq \alpha_1 < \dots < \alpha_n \leq Q} \prod_{l=1}^n \phi(x_{i_l}, w_{\alpha_l}) \phi(x_{i_{l+1}}, w_{\alpha_l}),
\end{equation}
with $i_{n+1} = i_1$.

For each fixed $\{x_i\}$, consider $h$ as a U-statistic with asymmetric kernel in the variables $\{w_\alpha\}$. The U-statistic kernel function is
\begin{equation}
h_{\text{kernel}}(\{w_{\alpha_l}\}) = \prod_{l=1}^n \phi(x_{i_l}, w_{\alpha_l}) \phi(x_{i_{l+1}}, w_{\alpha_l}).
\end{equation}

We first need to show that $h_{\text{kernel}}$ is absolutely integrable with respect to $\rho_\mathcal{W}^{\otimes n}$. Applying Hölder's inequality and using the square-integrability of $\phi$, we have
\begin{align*}
\int_{\mathcal{W}} d\rho_{\mathcal{W}}(w)\, \left| \phi(x_{i_l}, w) \phi(x_{i_{l+1}}, w) \right|  &\leq \left( \int_{\mathcal{W}} d\rho_{\mathcal{W}}(w) \, \phi^2(x_{i_l}, w) \right)^{1/2} \left( \int_{\mathcal{W}} d\rho_{\mathcal{W}}(w) \, \phi^2(x_{i_{l+1}}, w) \right)^{1/2} \\
&< \infty.
\end{align*}

Since the integrals are finite for all $x_{i_l}$ and $x_{i_{l+1}}$, the function $h_{\text{kernel}}$ is in $\mathcal{L}^1$.

By the strong law of large numbers for U-statistics with absolutely integrable asymmetric kernels \citep{janson2018renewal}, it follows that, as $Q \to \infty$,
\begin{equation}
h(\{w_\alpha\} \mid \{x_i\}) \xrightarrow{\text{a.s.}} g(\{x_i\}),
\end{equation}
where
\begin{equation}
g(\{x_i\}) = \frac{1}{\binom{P}{n}} \sum_{1 \leq i_1 < \dots < i_n \leq P} \prod_{l=1}^n k(x_{i_l}, x_{i_{l+1}}).
\end{equation}
Recall that $k$ is defined by
\begin{equation}
k(x, y) = \int_{\mathcal{W}} d\rho_{\mathcal{W}}(w) \, \phi(x, w) \phi(y, w) .
\end{equation}

Next, consider $g(\{x_i\})$ as a U-statistic over the variables $\{x_i\}$. The corresponding U-statistic kernel function is
\begin{equation}
g_{\text{kernel}}(\{x_{i_l}\}) = \prod_{l=1}^n k(x_{i_l}, x_{i_{l+1}}).
\end{equation}
Note $g_{\text{kernel}}$ is absolutely integrable. Applying the strong law of large numbers for U-statistics \citep{korolyuk2013theory}, it follows that, as $P \to \infty$,
\begin{equation}
g(\{x_i\}) \xrightarrow{\text{a.s.}} m(n),
\end{equation}
where
\begin{equation}
m(n) = \int \prod_{l=1}^n d\rho_\mathcal{X}(x_l) \, \prod_{l=1}^n k(x_l, x_{l+1})
\end{equation}
with $x_{n+1} = x_1$. Combining the two results for almost sure convergence and using the independence between $\{x_i\}$ and $\{w_\alpha\}$, we conclude that, as $P, Q \to \infty$,
\begin{equation}
\hat{m}(n) \xrightarrow{\text{a.s.}} m(n).
\end{equation}

\end{proof}
\section{Variance of \(\hat{m}\)}
In this section we sketch a derivation for an upper bound on the variance of our estimator \(\hat{m}(n)\) (see Theorem~\ref{varthe}). To motivate our derivation, consider a simpler setup. Suppose we have two sets
\begin{equation}
\mathcal{S}_{X} \coloneqq \{X_{j}\}_{j=1}^{A} \quad \text{and} \quad \mathcal{S}_{W} \coloneqq \{W_{j}\}_{j=1}^{B},
\end{equation}
with elements sampled from distributions \(\mu_{X}\) and \(\mu_{W}\), respectively. We assume that while elements within \(\mathcal{S}_{X}\) or \(\mathcal{S}_{W}\) may be correlated, the sets themselves are independent. Now define
\begin{equation}
V \coloneqq \frac{1}{AB}\sum_{i=1}^{A}\sum_{j=1}^{B} F(X_{i},W_{j}),
\end{equation}
where \(F:\mathcal{S}_{X}\times\mathcal{S}_{W}\to\mathbb{R}\). Our goal is to bound the variance of \(V\) by an expression of the form
\begin{equation}
\operatorname{Var}(V)\leq C\,\epsilon(A,B),
\end{equation}
with \(C\) an absolute constant and \(\epsilon(A,B)\) a function decreasing as \(A\) and \(B\) increase.

Writing the variance we have
\begin{equation}
\operatorname{Var}(V)=\left(\frac{1}{AB}\right)^{2}\sum_{i,j,k,l}\operatorname{Cov}\Bigl[F(X_{i},W_{j}),\,F(X_{k},W_{l})\Bigr].
\end{equation}
If we let \(G\) denote the number of nonzero covariance terms and note that \(\operatorname{Var}\bigl(F(X,W)\bigr)\) is the largest term, then
\begin{equation}
\operatorname{Var}(V)\leq \left(\frac{1}{AB}\right)^{2}G\,\operatorname{Var}\bigl(F(X,W)\bigr).
\end{equation}

In our estimator, each element \(X_{i}\) is a tuple \(\{x_{i_r}\}_{r=1}^{n}\) (with \(i_{1}<i_{2}<\cdots < i_{n}\)) and similarly each \(W_{l}\) is a tuple \(\{w_{l_r}\}_{r=1}^{n}\) (with \(l_{1}<l_{2}<\cdots <l_{n}\)). Thus, the set \(\mathcal{S}_{X}\times\mathcal{S}_{W}\) corresponds to all cyclic paths of length \(2n\) (with the cyclic constraint \(i_{n+1}=i_1\)). In our case, we define
\begin{equation}
F(X_i,W_l)\coloneqq\prod_{r=1}^{n}\phi(x_{i_r},w_{l_r})\,\phi(x_{i_{r+1}},w_{l_r}),
\end{equation}
where \(i_{n+1}=i_1\). Since the number of ways to choose an increasing sequence of indices is
\begin{equation}
A=\binom{P}{n}\quad \text{and}\quad B=\binom{Q}{n},
\end{equation}
we have \(V=\hat{m}(n)\). Moreover, one can show that the number of nonzero covariance terms is
\begin{equation}
G=\binom{P}{n}^2\binom{Q}{n}^2-\binom{P}{n}\binom{Q}{n}\binom{P-n}{n}\binom{Q-n}{n}.
\end{equation}
Thus,
\begin{equation}
\operatorname{Var}\bigl(\hat{m}(n)\bigr)\leq \Biggl(1-\frac{\binom{P-n}{n}\binom{Q-n}{n}}{\binom{P}{n}\binom{Q}{n}}\Biggr)
\operatorname{Var}\Biggl(\prod_{i=1}^{n}\phi(x_{i},w_{i})\,\phi(x_{i+1},w_{i})\Biggr).
\end{equation}

It remains to simplify the factor
\begin{equation}
1-\frac{\binom{P-n}{n}\binom{Q-n}{n}}{\binom{P}{n}\binom{Q}{n}}.
\end{equation}
Observe that
\begin{equation}
\frac{\binom{P-n}{n}}{\binom{P}{n}} = \prod_{i=0}^{n-1}\Bigl(1-\frac{n}{P-i}\Bigr)
\end{equation}
and similarly for \(Q\):
\begin{equation}
\frac{\binom{Q-n}{n}}{\binom{Q}{n}} = \prod_{i=0}^{n-1}\Bigl(1-\frac{n}{Q-i}\Bigr).
\end{equation}
Using \(1-x\leq e^{-x}\) (for \(x\geq 0\)) we obtain
\begin{equation}
\frac{\binom{P-n}{n}\binom{Q-n}{n}}{\binom{P}{n}\binom{Q}{n}}
\leq \exp\Bigl(-n^{2}\Bigl(\frac{1}{P}+\frac{1}{Q}\Bigr)\Bigr).
\end{equation}
Then, since \(1-e^{-x}\leq x\) for \(x\geq 0\),
\begin{equation}
1-\frac{\binom{P-n}{n}\binom{Q-n}{n}}{\binom{P}{n}\binom{Q}{n}}\leq n^{2}\Bigl(\frac{1}{P}+\frac{1}{Q}\Bigr).
\end{equation}
Thus, we finally obtain
\begin{equation}
\operatorname{Var}\bigl(\hat{m}(n)\bigr)\leq n^{2}\Bigl(\frac{1}{P}+\frac{1}{Q}\Bigr)
\,\operatorname{Var}\Biggl(\prod_{i=1}^{n}\phi(x_{i},w_{i})\,\phi(x_{i+1},w_{i})\Biggr).
\end{equation}

\section{The error of the eigenvalue recovery}
In \cite{kong2017} the following relationship is established between the expected total absolute error in the recovered eigenvalues and the variance of a moment estimator. Specifically, if one recovers \(d\) eigenvalues \(\{\hat{\lambda}_i\}_{i=1}^{d}\) from the estimated moments \(\{\hat{m}(n)\}_{n=1}^{k}\) (using, for example, a moment–to–spectrum algorithm), then
\begin{equation}
    \Biggl\langle \sum_{i=1}^{d}\Bigl|\lambda_{i}-\hat{\lambda}_{i}\Bigr| \Biggr\rangle 
    \leq bd\Biggl(C'3^{k}k\Bigl(\sqrt{\operatorname{Var}\bigl(\hat{m}(k)\bigr)}+dk\epsilon\Bigr)
    +\frac{C}{k}+\frac{1}{d}\Biggr),
\end{equation}
where \(C\), \(C'\), and \(\epsilon\) are positive constants (with \(\epsilon\) capturing additional approximation error due to discretization of the spectral density), \(k\) denotes the order of the moment used in the reconstruction, $b$ is the upper bound of the eigenvalues, and \(d\) is the (finite) rank of the operator. By substituting the variance bound derived above into this inequality, one obtains an explicit upper bound on the expected total absolute error in the recovered eigenvalues.



\section{Spectrum of the general RBF kernel with Gaussian input}

\subsection{Derivation of spectral moments}

We study the kernel operator for the radial basis function (RBF)
kernel
\begin{equation}
k(x,x')=e^{-\frac{1}{2}\left(x-x'\right)^{\top}\Sigma^{-1}\left(x-x'\right)}
\end{equation}
and input distribution $\rho_{\mathcal{X}}=\mathcal{N}(0,\Sigma_{x})$.

Then the $n$-th spectral moment is moment is given by
\begin{align}
m(n)=&\int\prod_{i=1}^{n}d\rho_{\mathcal{X}}(x_{i})\:\prod_{i=1}^{n}k(x_{i},x_{i+1})
\\
=&\left(\left(2\pi\right)^{d}|\Sigma_{x}|\right)^{-n/2}\int\prod_{i=1}^{n}dx_{i}\:\exp-\frac{1}{2}\left(\sum_{i=1}^{n-1}\left(x_{i}-x_{i+1}\right)^{\top}\Sigma^{-1}\left(x_{i}-x_{i+1}\right)+\sum_{i=1}^{n}x_{i}^{\top}\Sigma_{x}^{-1}x_{i}\right)
\end{align}

where $x_{n+1}=x_{1}$. The integrand simplifies to
\begin{equation}
\exp-\frac{1}{2}\left[\sum_{i=1}^{n}x_{i}\left(2\Sigma^{-1}+\Sigma_{x}^{-1}\right)x_{i}-2\left(\sum_{i=1}^{n-1}x_{i}\Sigma^{-1}x_{i+1}+x_{1}\Sigma^{-1}x_{n}\right)\right],
\end{equation}
which can be written as

\begin{equation}
\exp-\frac{1}{2}\bar{x}_{n}^{\top}\begin{pmatrix}D & -\Sigma^{-1} & 0 & \cdots & 0 & 0 & -\Sigma^{-1}\\
-\Sigma^{-1} & D & -\Sigma^{-1} & \cdots & 0 & 0 & 0\\
0 & -\Sigma^{-1} & D & \cdots & 0 & 0 & 0\\
\vdots & \vdots & \vdots & \ddots & \vdots & \vdots & \vdots\\
0 & 0 & 0 & \cdots & D & -\Sigma^{-1} & 0\\
0 & 0 & 0 & \cdots & -\Sigma^{-1} & D & -\Sigma^{-1}\\
-\Sigma^{-1} & 0 & 0 & \cdots & 0 & -\Sigma^{-1} & D
\end{pmatrix}\bar{x}_{n}
\end{equation}
where $D\coloneqq2\Sigma^{-1}+\Sigma_{x}^{-1}$, and $\bar{x}_{n}\coloneqq\left[x_{1}^\top,\ldots,x_{n}^\top\right]^\top\in\mathbb{R}^{nd}$.
Denoting the above block matrix as $M$, we have simplified the moment
equation to
\begin{equation}
m(n)=\left(\left(2\pi\right)^{d}|\Sigma_{x}|\right)^{-n/2}\int d\bar{x}_{n}\:\exp\left(-\frac{1}{2}\bar{x}_{n}^{\top}M\bar{x}_{n}\right).
\end{equation}
Solving the Gaussian integral gives
\begin{equation}
m(n)=\left(\det\Sigma_{x}^{n}\det M\right)^{-\frac{1}{2}}.
\end{equation}
In general, the determinant of a block circulant matrix:
\begin{equation}
M=\begin{pmatrix}R_{0} & R_{1} & \cdots & R_{n-1}\\
R_{n-1} & R_{0} & \cdots & R_{n-2}\\
\vdots & \vdots & \ddots & \vdots\\
R_{1} & R_{2} & \cdots & R_{0}
\end{pmatrix},
\end{equation}
is given by
\begin{equation}
\det M=\prod_{q=0}^{n-1}\det\left(\sum_{l=0}^{n-1}e^{2\pi iql/n}R_{l}\right)
\end{equation}
After some algebra, the determinant of the block matrix becomes
\begin{equation}
\det M=\prod_{q=1}^{n}\det\left\{ 2\left[1-\cos\left(2\pi\frac{q}{n}\right)\right]\Sigma^{-1}+\Sigma_{x}^{-1}\right\} .
\end{equation}
Therefore,
\begin{equation}
m(n)=\left(\prod_{q=1}^{n}\det\left\{ 2\left[1-\cos\left(2\pi\frac{q}{n}\right)\right]\Sigma^{-1}\Sigma_{x}+I\right\} \right)^{-\frac{1}{2}}.
\end{equation}
We can simplify further. Let $\left\{ \eta_{i}\right\} _{i=1}^{d}$
be the eigenvalues of $\Sigma^{-1}\Sigma_{x}$, which are the same
as the eigenvalues of $\Sigma_{x}\Sigma^{-1}$. Note that $\eta_{i}$
is always real, since $\Sigma^{-1}$ and $\Sigma_{x}$ are positive
semi-definite, and the product of two positive semi-definite matrices
has real eigenvalues. Then,
\begin{equation}
m(n)=\prod_{q=1}^{n}\prod_{i=1}^{d}\left[1+2\eta_{i}-2\eta_{i}\cos\left(2\pi\frac{q}{n}\right)\right]^{-\frac{1}{2}},
\end{equation}
which can be written using the law of cosines:
\begin{equation}
m(n)=\prod_{q=1}^{n}\prod_{i=1}^{d}\frac{1}{\eta_{i}}\left[\left(\frac{1+\sqrt{1+4\eta_{i}}}{2\eta_{i}}\right)^{2}+\left(\frac{-1+\sqrt{1+4\eta_{i}}}{2\eta_{i}}\right)^{2}-\frac{2}{\eta_{i}}\cos\left(2\pi\frac{q}{n}\right)\right]^{-\frac{1}{2}}
\end{equation}
Let us define $\phi_{\eta_{i}}=\frac{1+\sqrt{1+4\eta_{i}}}{2\eta_{i}}$, so that
$\eta_{i}^{-1}\phi_{\eta_{i}}^{-1}=\frac{-1+\sqrt{1+4\eta_{i}}}{2\eta_{i}}$.
Using these definitions, we can rewrite the above as
\begin{equation}
m(n)=\prod_{i=1}^{d}\phi_{\eta_{i}}^{n}\left\{ \prod_{q=1}^{n}\left[\left(\eta_{i}\phi_{\eta_{i}}^{2}\right)^{2}-2\cos\left(2\pi\frac{q}{n}\right)\left(\eta_{i}\phi_{\eta_{i}}^{2}\right)+1\right]\right\} ^{-\frac{1}{2}}
\end{equation}
Now, the following identity can be used: $\prod_{q=1}^{n}\left(x^{2}-2\cos\left(2\pi\frac{q}{n}\right)x+1\right)=\left(x^{n}-1\right)^{2}$.
Therefore, we arrive at
\begin{equation}
m(n)=\prod_{i=1}^{d}\frac{1}{\eta_{i}^{n}\phi_{\eta_{i}}^{n}-\phi_{\eta_{i}}^{-n}}.
\end{equation}

\subsection{Derivation of eigenvalues}

Consider $f_{u}(x)=\prod_{i=1}^{d}H_{u_{i}}(x_{i})e^{-\alpha_{i}x_{i}^{2}}$
where $H_{c}(x)$ is a $c$-th order polynomial with leading
coefficient of $1$. $u\coloneqq\left\{ u_{i}\right\} _{i=1}^{d}$ is
a multiset of $d$ natural numbers. We require that $T_{k}f_{u}=\lambda_{u}f_{u}$,
to solve for the eigenvalues and eigenfunctions. Let us first find
an expression for $T_{k}f_{u}$:

\begin{equation}
\left[T_{k}f_{u}\right](y)=\frac{1}{\sqrt{\left(2\pi\right)^{d}\prod_{i=1}^{d}\eta_{i}}}e^{-\frac{1}{2}\sum_{i=1}^{d}y_{i}^{2}}\int\prod_{i=1}^{d}dx_{i}\:\prod_{i=1}^{d}H_{u_{i}}(x_{i})\exp\left[\sum_{i=1}^{d}-\left(\frac{1}{2\eta_{i}}+\frac{1}{2}+\alpha_{i}\right)x_{i}^{2}+x_{i}y_{i}\right]
\end{equation}
After some algebra, we arrive at 
\begin{equation}
\left[T_{k}f_{u}\right](y)=\sqrt{\prod_{i=1}^{d}\frac{1}{2\alpha_{i}'\eta_{i}}}\left\langle \prod_{i=1}^{d}H_{u_{i}}(x_{i})\right\rangle _{{\mathcal N}}e^{-\sum_{i=1}^{d}\left(\frac{1}{2}-\frac{1}{4\alpha_{i}'}\right)y_{i}^{2}}
\end{equation}
\begin{equation}
{\mathcal N}\left(\left\{ \frac{1}{2\alpha_{i}'}y_{i}\right\} _{i=1}^{d},\text{diag}\left(\left\{ \frac{1}{2\alpha_{i}'}\right\} _{i=1}^{d}\right)\right)
\end{equation}
where $\alpha'_{i}=\frac{1}{2\eta_{i}}+\frac{1}{2}+\alpha_{i}$, and
$\text{diag}\left(\left\{ a_{i}\right\} _{i=1}^{d}\right)$ is a diagonal
matrix whose $i$-th diagonal entry is $a_{i}$.

Now the eigenvalue equation $\left[T_{k}f_{u}\right](y)=\lambda_{u}f_{u}(y)$
requires that
\begin{equation}
\sqrt{\prod_{i=1}^{d}\frac{1}{2\alpha_{i}'\eta_{i}}}\left\langle \prod_{i=1}^{d}H_{u_{i}}(x_{i})\right\rangle _{{\mathcal N}}e^{-\sum_{i=1}^{d}\left(\frac{1}{2}-\frac{1}{4\alpha_{i}'}\right)y_{i}^{2}}=\lambda_{u}\prod_{i=1}^{d}H_{u_{i}}(y_{i})e^{-\sum_{i=1}^{d}\alpha_{i}y_{i}^{2}}
\end{equation}
Equating the exponents $\frac{1}{2}-\frac{1}{4\alpha_{i}'}=\alpha_{i}$
gives 
\begin{equation}
\alpha_{i}=\frac{-1+\sqrt{1+4\eta_{i}}}{4\eta_{i}}
\end{equation}
Recall $\eta_{i}^{-1}\phi_{\eta_{i}}^{-1}=\frac{-1+\sqrt{1+4\eta_{i}}}{2\eta_{i}}$.
Therefore $\sqrt{\frac{1}{2\alpha'\eta_{i}}}=\eta_{i}^{-1}\phi_{\eta_{i}}^{-1}$,
which means the above eigenvalue equation simplifies:

\begin{equation}
\left(\prod_{i=1}^{d}\eta_{i}^{-1}\phi_{\eta_{i}}^{-1}\right)\left\langle \prod_{i=1}^{d}H_{u_{i}}(x_{i})\right\rangle _{{\mathcal N}}=\lambda_{u}\prod_{i=1}^{d}H_{u_{i}}(y_{i})
\end{equation}

\begin{equation}
{\mathcal N}\left(\left\{ \eta_{i}^{-1}\phi_{\eta_{i}}^{-2}y_{i}\right\} ,\text{diag}\left(\left\{ \eta_{i}^{-1}\phi_{\eta_{i}}^{-2}\right\} \right)\right).
\end{equation}
Since each dimension is independent,

\begin{equation}
\left(\prod_{i=1}^{d}\eta_{i}^{-1}\phi_{\eta_{i}}^{-1}\right)\prod_{i=1}^{d}\left\langle H_{u_{i}}(x_{i})\right\rangle _{{\mathcal N}}=\lambda_{u}\prod_{i=1}^{d}H_{u_{i}}(y_{i}).
\end{equation}
Consider an example polynomial:
\begin{equation}
\left\langle H_{u_{i}}(x_{i})\right\rangle _{{\mathcal N}}=\left\langle x_{i}^{u_{i}}+{\mathcal O}(x_{i}^{u_{i}-1})\right\rangle _{{\mathcal N}}=\left(\eta_{i}^{-1}\phi_{\eta_{i}}^{-2}\right)^{u_{i}}y^{u_{i}}+{\mathcal O}(y^{u_{i}-1}).
\end{equation}
Showing only the leading order terms of the LHS and RHS of the eigenvalue equation, we see
\begin{equation}
\left(\prod_{i=1}^{d}\eta_{i}^{-1}\phi_{\eta_{i}}^{-1}\right)\left(\eta_{i}^{-1}\phi_{\eta_{i}}^{-2}\right)^{u_{i}}y^{u_{i}}+{\mathcal O}(y^{u_{i}-1})=\lambda_{u}y^{u_{i}}+{\mathcal O}(y^{u_{i}-1}).
\end{equation}
Equating the coefficients of the leading order terms, we find that
\begin{equation}
\lambda_{u}=\prod_{i=1}^{d}\left(\eta_{i}^{1+u_{i}}\phi_{\eta_{i}}^{1+2u_{i}}\right)^{-1}.
\end{equation}
The same eigenvalues can be found via the Taylor series expansion
of $m(n)$. Let us define $r_{i}=\eta_{i}^{-1}\phi_{\eta_{i}}^{-2}$
and $s=\prod_{i=1}^{d}\left(\eta_{i}^{-1}\phi_{\eta_{i}}^{-1}\right)$.
Then,
\begin{align}
m(n)=&s^{n}\prod_{i=1}^{d}\frac{1}{1-r_{i}^{n}}=s^{n}\prod_{i=1}^{d}\left(1+r_{i}^{n}+r_{i}^{2n}+\cdots\right)
\\
=&s^{n}\left(1+r_{1}^{n}+r_{1}^{2n}+\cdots\right)\left(1+r_{2}^{n}+r_{2}^{2n}+\cdots\right)\left(1+r_{3}^{n}+r_{3}^{2n}+\cdots\right).
\end{align}
Notice that expanding the above product yields a sum of terms of $n$-th degree: $\left(sr_{1}^{a}r_{2}^{b}r_{3}^{c}r_{4}^{d}\cdots\right)^n$
where $\{a,b,c,d,\ldots\}$ is a multiset of integers. This implies
that
\begin{equation}
\lambda_{u}=s\prod_{i=1}^{d}r_{i}^{u_{i}}.
\end{equation}
Plugging in the definitions for $r_{i}$ and $s$, we get
\begin{equation}
\lambda_{u}=\prod_{i=1}^{d}\left(\eta_{i}^{1+u_{i}}\phi_{\eta_{i}}^{1+2u_{i}}\right)^{-1}
\end{equation}
which agrees with the results of the earlier derivation.

\section{Bias and variance of kernel integral operator moment estimators (empirical)}

In the numerical experiments with the radial basis function (RBF) kernel and Gaussian input distribution, we observe that our estimator achieves both the lowest bias and variance error, across all configurations that are tested.

\begin{figure*}[ht]
    \centering
    \includegraphics[width=\textwidth]{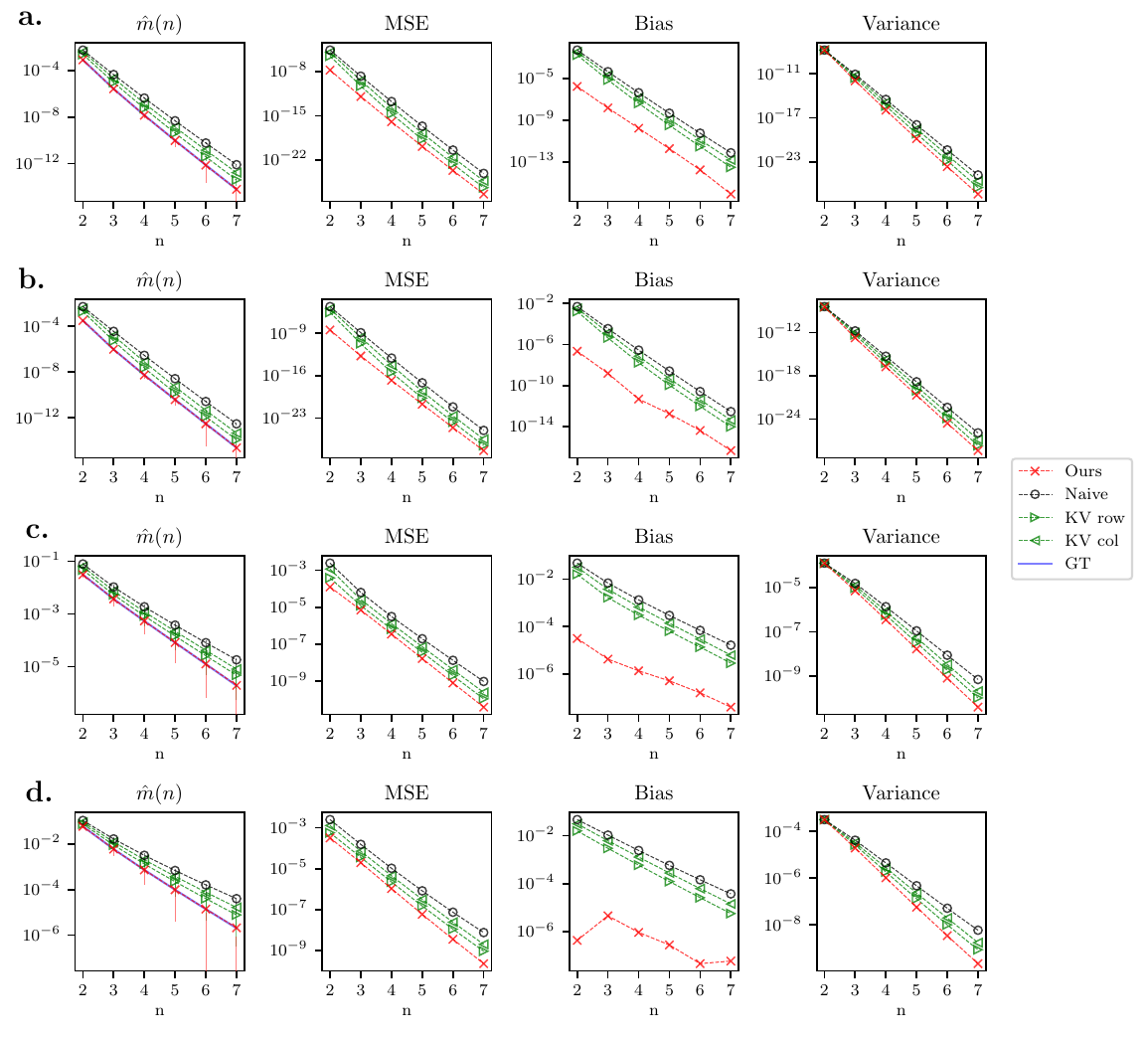}
    \caption{Performance of the estimators with the RBF kernel. Columns from left to right: the estimated moments; the mean-square error between the estimated moments and the ground true moments averaged over multiple samples of $\Phi$'s  ($\left<\hat{m}(n)-m(n)\right>_\Phi$); bias error ($ \left<\hat{m}(n)\right> - m(n)$); variance error ($\left<\hat{m}(n)^2\right> - \left<\hat{m}(n)\right>^2$). \textbf{a.} 
    $P=300$, $Q=600$, $d=5$, $\Sigma_x=I_{d\times d}$, $\Sigma=0.25 I_{d\times d}$. \textbf{b.} $P=300$, $Q=600$, $d=10$, $\Sigma_x=I_{d\times d}$, $\Sigma=I_{d\times d}$. \textbf{c.} $P=30$, $Q=60$, $d=10$, $\Sigma_x=I_{d\times d}$, $\Sigma=4 I_{d\times d}$. \textbf{d.} $P=30$, $Q=60$, $d=4$, $\Sigma_x=I_{d\times d}$, $\Sigma=0.25 I_{d\times d}$.}
    \label{fig:rbf_config}
\end{figure*}

Figure \ref{fig:rbf_config}) shows the performance of the different estimators along with a detailed breakdown of the error in terms of the experimentally observed bias and variance.

\section{Moment estimation with noise}



An unbiased estimate of the $m(n)$ can be obtained even when the measurements are corrupted by correlated noise by utilizing measurements over two or more trials.
In these experiments, we add noise to the measurement matrix resulting in both row-correlated and column-correlated noise,
\begin{equation}
\left< \Phi_{i\alpha}^{(t)}\Phi_{j\beta}^{(t)} \right>_{t} \propto \delta_{ij}+\delta_{\alpha \beta}
\end{equation}
where $t$ is the trial index, and $\Phi_{i\alpha}^{(t)}$ is centered.
As noted in the main text, the following product with alternation between two trials gives an unbiased estimate of $m(n)$ in the presence of the correlated noise:
\begin{equation}\label{eq:T2case}
\hat{m}'_\text{alt-\{1,2\}}(n)=\prod_{l=1}^{n}\Phi^{(1)}_{i_{l}\alpha_{l}}\Phi^{(2)}_{i_{l+1}\alpha_{l}}
\end{equation}
with the trace constraint $i_{n+1}=i_1$. Algorithm \ref{algo_alt_2} details how to compute the estimator with two trial measurements. In general, when there are $T$ total trials, we can write
\begin{equation}\label{eq:altT}
\hat{m}'_{\text{alt-}\mathcal{T}}(n)=\prod_{l=1}^{n}\Phi^{(t_{2l-1})}_{i_{l}\alpha_{l}}\Phi^{(t_{2l})}_{i_{l+1}\alpha_{l}}.
\end{equation}
$\mathcal{T}=\{t_l\}^{2n}_{l=1}$ is a ordered multiset with cardinality $2n$ of the trial index set $\{1,2,\ldots,T\}$. The necessary and sufficient condition for \eqref{eq:altT} to be an unbiased estimator in the presence of correlated noise is that for all $i$ and $j$ where $\left|i-j\right|=1$, and for $i=1$ and $j=2n$, $t_i$ and $t_j$ take distinct values. The algorithm for this case is presented in Algorithm \ref{algo_alt_T}.
Alternatively, one can use algorithm \ref{algo_alt_2} for \eqref{eq:T2case} by randomly selecting two distinct trials from $\{1,\ldots,T\}$ repeatedly and then averaging over the resulting estimates.

\begin{algorithm}[t!]
\caption{Computation of $\hat{m}_\text{alt}(n)$ for $n = 2$ to $n_{\text{max}}$ when $T=2$}
\begin{algorithmic}[1]
\Require $\Phi^{(1)},\Phi^{(2)} \in \mathbb{R}^{P \times Q}$, $n_{\text{max}}$
\For{$h \leftarrow 1$ \textbf{to} $P$}
    \State Initialize $S$ as a $P \times Q$ zero matrix.
    \State Set $S_{hi} \leftarrow P \Phi_{hi}^{(1)}$  $\forall i \in [1, Q]$
    \For{$n \leftarrow 2$ \textbf{to} $n_{\text{max}}$}

        \State Update
        $S_{ab} \leftarrow 
        \dfrac{n^{2}\sum_{l=h+n-2}^{a-1} \sum_{k=n-1}^{b-1}  S_{lk} \Phi^{(2)}_{ak} \Phi^{(1)}_{ab}}{(P - n + 1)(Q - n + 1)}$  $\forall a \in [h+n-1, P]$, $\forall b \in [n, Q]$.
        \label{algo_alt_2:update_alt}
        
        \State 
        Compute  $\hat{m}_\text{alt}^{(h)}(n) \leftarrow \dfrac{1}{PQ} \sum_{i=h+n-1}^{P} \sum_{j=n}^{Q} S_{ij} \Phi_{hj}^{(2)}$.
    \EndFor
\EndFor
\State Get $\hat{m}_\text{alt}(n) \gets \dfrac{1}{P} \displaystyle\sum_{h=1}^{P-n+1} \hat{m}_\text{alt}^{(h)}(n)$ $\forall n \in [2,n_\text{max}]$.
\end{algorithmic}
\label{algo_alt_2}
\end{algorithm}

\begin{algorithm}[t!]
\caption{Computation of $\hat{m}_\text{alt}(n)$ for $n = 2$ to $n_{\text{max}}$ for arbitrary $T\geq 2$}
\begin{algorithmic}[1]
\Require $\{\Phi^{(t)}\}_{t=1}^{T} \in \mathbb{R}^{P \times Q}$, $n_{\text{max}}$, 
\For{$h \leftarrow 1$ \textbf{to} $P$}
    \State Initialize $S$ as a $P \times Q$ zero matrix.
    \State Set $S_{hi} \leftarrow P \Phi_{hi}^{(t_1)}$  $\forall i \in [1, Q]$.
    \State Choose $t_1\in\{1,\ldots,T\}$
    \For{$n \leftarrow 2$ \textbf{to} $n_{\text{max}}$}
        \State Choose 
        $t_{2n-2},t_{2n-1}\in\{1,\ldots,T\}$ such that $t_{2n-3}\neq t_{2n-2}$ and $t_{2n-2} \neq t_{2n-1}$.
        \State Update
        $S_{ab} \leftarrow 
        \dfrac{n^{2}\sum_{l=h+n-2}^{a-1} \sum_{k=n-1}^{b-1}  S_{lk} \Phi^{(t_{2n-2})}_{ak} \Phi^{(t_{2n-1})}_{ab}}{(P - n + 1)(Q - n + 1)}$  $\forall a \in [h+n-1, P]$, $\forall b \in [n, Q]$.
        \label{algo_alt_T:update_alt}

        \State Choose $t_r$ such that $t_r\neq t_{2n-1}$ and $t_r\neq t_1$.
        \State 
        Compute  $\hat{m}_\text{alt}^{(h)}(n) \leftarrow \dfrac{1}{PQ} \sum_{i=h+n-1}^{P} \sum_{j=n}^{Q} S_{ij} \Phi_{hj}^{(t_{r})}$.
    \EndFor
\EndFor
\State Get $\hat{m}_\text{alt}(n) \gets \dfrac{1}{P} \displaystyle\sum_{h=1}^{P-n+1} \hat{m}_\text{alt}^{(h)}(n)$ $\forall n \in [2,n_\text{max}]$.
\end{algorithmic}
\label{algo_alt_T}
\end{algorithm}

\subsection{Numerical estimation results for noisy measurements}

\begin{figure*}[ht]
    \centering
    \includegraphics[width=\textwidth]{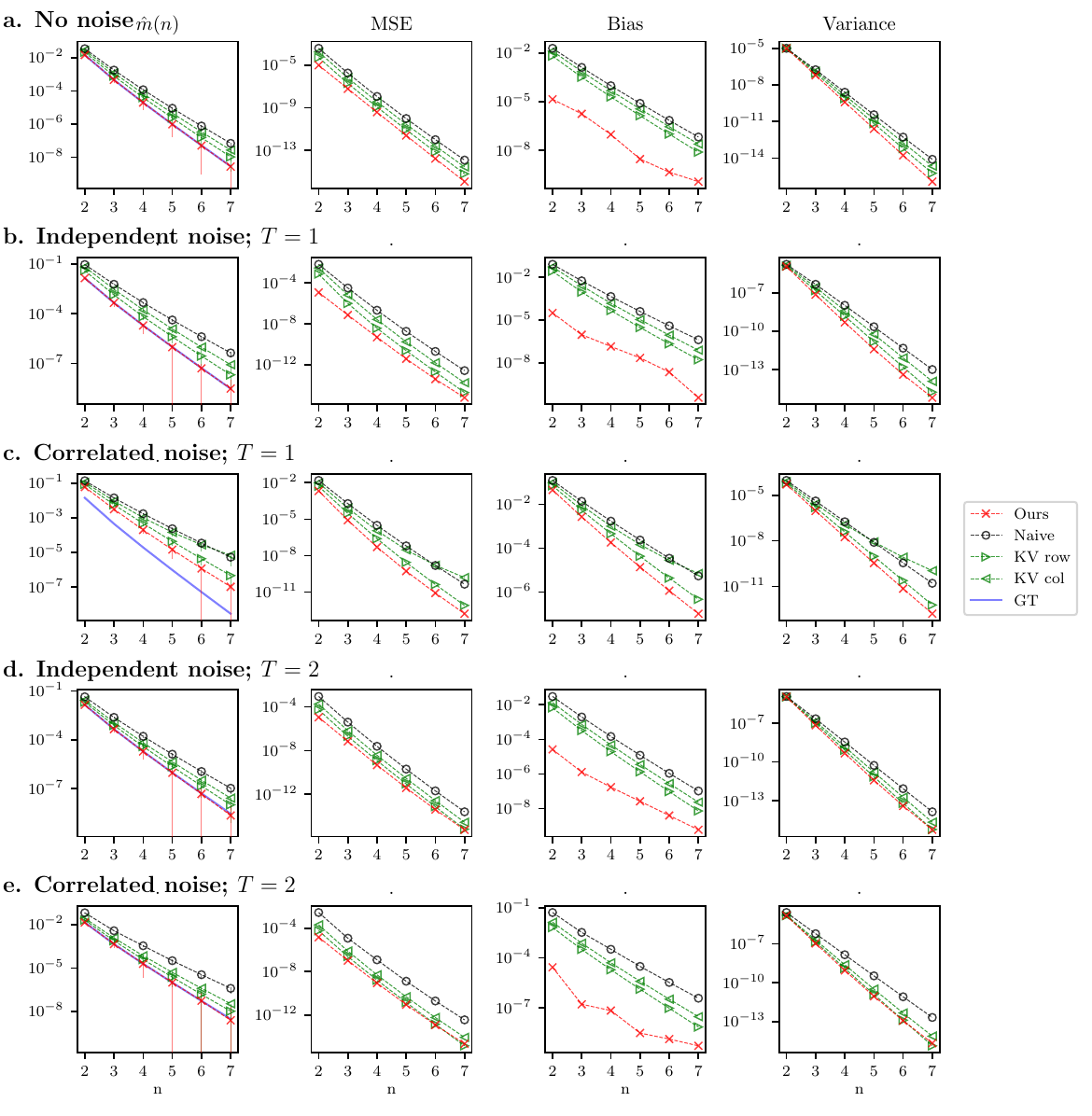}
    \caption{Performance of the estimators in the presence of independent or correlated noise for the RBF kernel. $P=75$, $Q=15$, $d=3$, $\Sigma_x=I_{d\times d}$, $\Sigma=0.25 I_{d\times d}$. Columns from left to right: the estimated moments; the mean-square error between the estimated moments and the ground true moments averaged over multiple samples of $\Phi$'s  ($\left<\hat{m}(n)-m(n)\right>_\Phi$); bias error ($ \left<\hat{m}(n)\right> - m(n)$); variance error ($\left<\hat{m}(n)^2\right> - \left<\hat{m}(n)\right>^2$). \textbf{a.} No noise case. \textbf{b.} The data is corrupted by an additive independent noise sampled from the standard normal distribution. The number of trials is $T=1$. \textbf{c.} The data is corrupted by an additive correlated noise (not independently) sampled from the standard normal distribution. For a given input $i$, the noise is correlated between entry $\Phi_{i\alpha}$ and $\Phi_{i\beta}$ for $|\alpha-\beta|>10$. \textbf{d.} Estimators alternating between measurements from two trials with independent noise. \textbf{e.} Estimators alternating between measurements from two trials with correlated noise.}
    \label{fig:noisy_rbf}
\end{figure*}

The cross-trial alternation method can also be applied to the naive as well as \cite{kong2017} estimators to remove the effects of independent and correlated noise. We test the estimators with or without the cross-trial alternation, on the RBF kernel measurement data with no noise, independent noise, and correlated noise (Figure \ref{fig:noisy_rbf}). These numerical tests confirm that with only two trials $T=2$, our estimator is still unbiased even when the measurement matrix is corrupted by correlated noise (Figure \ref{fig:noisy_rbf}e). It can also be seen that for independent noise, we only need one trial ($T=1$) of the measurement matrix to obtain an unbiased estimate (Figure \ref{fig:noisy_rbf}b). Our estimator achieves the lowest bias and variance error across nearly all configurations.

\section{ReLU network learning implementation}


The feature learning example in the main text is demonstrated with a single-hidden layer neural network. Each input $x\in\mathbb{R}^d$ is a flattened vector of Fashion-MNIST image's $28\times 28$ pixels. The $w_i\in\mathbb{R}^d$ represents the weight vector for the $i$-th neuron in the hidden layer with a total $N$ number of neurons. Let $a_j\in\mathbb{R}^N$ be the weight vector for the $j$-th neuron in the output layer with 10 neurons, each corresponding to one class in the Fashion-MNIST dataset. Explicitly, the value of the $j$-th output neuron can be written as:
\begin{equation}
    y_j = \sum_{i=1}^N \phi(x,w_i) a_{ji}
\end{equation}
where $a_{ji}\in\mathbb{R}$ is an $i$-th element of the vector $a_j \in \mathbb{R}^N$, and 
\begin{equation}
    \phi(x,w)\coloneqq \max\left(x\cdot w,0\right).
\end{equation}
We minimize the sum of the square of the difference between the 10-dimensional network output and the 10-dimensional one-hot vector that indicates the class membership. We use 10,000 images for training the network.

The weights are initialized according to the maximal update parameterization ($\mu P$) to ensure feature learning even when $N$ is large \citep{yang2022tensor}. For the single-hidden layer neural network that will be trained with Adam-SGD, the $\mu P$ initialization is:
\begin{equation}
    w_i\sim\mathcal{N}\left(0,\frac{1}{N}I_{d\times d}\right)
\end{equation}
\begin{equation}
    a_j\sim\mathcal{N}\left(0,\frac{1}{N}I_{N\times N}\right)
\end{equation}
with independent sampling across all $i$'s and $j$'s. For proper gradient scaling during the backward pass, $\mu P$ requires the following modification to the model:
\begin{equation}
    y_j = \sum_{i=1}^N \frac{1}{\sqrt{N}}\phi(x,w_i) a_{ji}
\end{equation}
\begin{equation}
    \phi(x,w)\coloneqq \sqrt{N}\max\left(x\cdot w,0\right).
\end{equation}
which does not affect the forward pass. The learning rates are fixed to $\frac{.01}{N}$. Note that this scaling is specific for Adam-SGD \citep{yang2022tensor}. We randomly sample 32 images from 10,000 training images for a mini-batch, and train each network for 120 epochs.

In the main text, we train networks with the following widths ($N$): 32, 64, 128, 256, 512, and 1024, and compute the spectral moments every even epoch. Each entry of the measurement matrix $\Phi\in\mathbb{R}^{P\times Q}$ in this case is obtained as:
\begin{equation}
    \Phi_{i\alpha}=\frac{\phi(x_i,w_\alpha)}{\sqrt{\frac{1}{PQ}\sum_{j=1}^P \sum_{\beta=1}^Q \phi(x_j,w_\beta)^2 } }
\end{equation}
for $P=1000$ number of test images $\{x_i\}_{i=1}^P$, and $Q$ number of neurons $\{w_\alpha\}_{\alpha=1}^Q$. In the main text, we showed the results when measuring all neurons $Q=N$ in the hidden layer. For all training, Quadro GV100 GPU is used.

Next, we explore the case where the measurement matrix consists of only partial observations of the neurons $Q<N$.

\subsection{Estimation from partial measurements}

\begin{figure*}[ht]
    \centering
    \includegraphics[width=\textwidth]{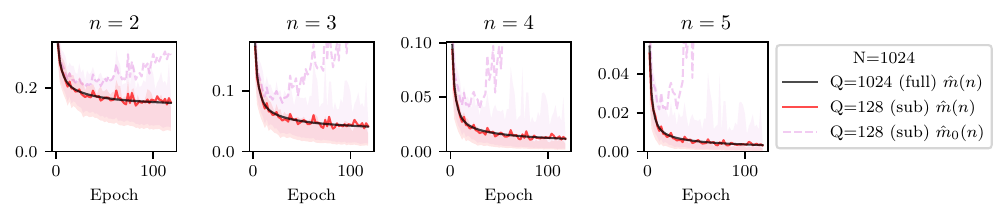}
    \caption{Single hidden layer neural network with $N=1024$ neurons in the hidden layer. Each plot corresponds to a different moment order $n$. $P=1000$ test images are used. Black line: mean value of our estimator $\hat{m}(n)$ applied to $\Phi$ with all neurons $Q=N$. Solid red line: mean value of our estimator $\hat{m}(n)$ applied to $\Phi$ with subsampled neurons $Q=128<N$. Dotted magenta line: mean value of the naive estimator $\hat{m}_0(n)$ applied to $\Phi$ with subsampled neurons. The shaded regions indicate a 50\% confidence interval. For the naive estimator, the mean value falls outside the confidence interval.}
    \label{fig:train_supp1}
\end{figure*}

\begin{figure*}[ht]
    \centering
    \includegraphics[width=\textwidth]{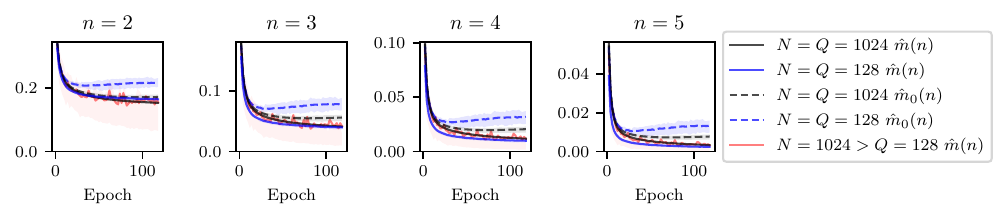}
    \caption{Single hidden layer neural networks trained with $N=1024$ (black) and $N=128$ (blue) hidden layer neurons. Each plot corresponds to a different moment order $n$. Each measurement matrix observes all neurons $Q=N$, except for one case (red) where $Q=128$ neurons are randomly subsampled from $N=1024$ neurons. $P=1000$ test images are used.  The solid lines are the values of our estimator $\hat{m}(n)$, and the dotted lines are the values of the naive estimator $\hat{m}_0(n)$. The shaded regions indicate a 50\% confidence interval.}
    \label{fig:train_supp2}
\end{figure*}

Here we explore the following question: given a wide neural network (large $N$), how closely do the moment estimates from observing all neurons in the hidden layer $Q=N$ and the estimates from observing random subsamples of the neurons $Q<N$ match?

To check this, for each epoch of training a wide single-hidden layer neural network with $N=1024$ hidden neurons using the above specifications, we estimate the moments from a measurement matrix $\Phi_\text{all}$ from all neurons $Q=N$ and the moments from another measurement matrix $\Phi_\text{sub}$ that observes a smaller set of neurons $Q=128<N$. The numerical results show that in every epoch, these two estimates match very closely (Figure \ref{fig:train_supp1}). This means that we do not need to observe all neurons in a neural network to determine the spectral properties of the kernel operator. This can be particularly useful for large-scale networks, such as in state-of-the-art Transformer models, where computing the covariance matrix is highly memory and computationally intensive.

We also compare these moments to those from networks trained with smaller sizes ($N=128$), and find that the resulting kernel moments estimates also match (Figure \ref{fig:train_supp2}), as expected based upon the results from the main text.

\section{Code availability}
The code for the estimators and generating all figures is publicly available on \href{https://github.com/badooki/spectral_moments}{Github}.

\end{document}